# Dynamic Predictions of Postoperative Complications from Explainable, Uncertainty-Aware, and Multi-Task Deep Neural Networks

Benjamin Shickel, PhD [1,6], Tyler J. Loftus, MD [2,6], Matthew Ruppert, BS [1,3,6], Gilbert R. Upchurch Jr., MD [2], Tezcan Ozrazgat-Baslanti, PhD [1,3,6], Parisa Rashidi, PhD [1,4,5,6], Azra Bihorac, MD, MS [1,3,6,*]

[1] Department of Medicine, University of Florida, Gainesville, FL, 32611, USA
[2] Department of Surgery, University of Florida, Gainesville, FL, 32611, USA
[3] Precision and Intelligent Systems in Medicine (PRISMAp), University of Florida, Gainesville, FL, 32611, USA
[4] Department of Biomedical Engineering, University of Florida, Gainesville, FL, 32611, USA
[5] Intelligent Health Lab (i-Heal), University of Florida, Gainesville, FL, 32611, USA
[6] Intelligent Critical Care Center (IC[3]), University of Florida, Gainesville, FL, 32611, USA

Correspondence*:
Azra Bihorac, MD MS FCCM FASN
Senior Associate Dean for Research Affairs
R. Glenn Davis Associate Professor of Medicine, Surgery and Anesthesiology
Director, Intelligent Critical Care Center
University of Florida Health, Gainesville, Florida

Address: PO Box 100224, Gainesville, FL 32610-0254
Office: (352) 273-5995
Center: (352)-294-8509
Email: abihorac@ufl.edu

## ABSTRACT

Accurate prediction of postoperative complications can inform shared decisions regarding prognosis, preoperative risk-reduction, and postoperative resource use. We hypothesized that multi-task deep learning models would outperform random forest models in predicting postoperative complications, and that integrating high-resolution intraoperative physiological time series would result in more granular and personalized health representations that would improve prognostication compared to preoperative predictions. In a longitudinal cohort study of 56,242 patients undergoing 67,481 inpatient surgical procedures at a university medical center, we compared deep learning models with random forests for predicting nine common postoperative complications using preoperative, intraoperative, and perioperative patient data. Our study indicated several significant results across experimental settings that suggest the utility of deep learning for capturing more precise representations of patient health for augmented surgical decision support. Multi-task learning improved efficiency by reducing computational resources without compromising predictive performance. Integrated gradients interpretability mechanisms identified potentially modifiable risk factors for each complication. Monte Carlo dropout methods provided a quantitative measure of prediction uncertainty that has the potential to enhance clinical trust. Multi-task learning, interpretability mechanisms, and uncertainty metrics demonstrated potential to facilitate effective clinical implementation.

## INTRODUCTION

In the United States, more than 15 million major, inpatient surgeries are performed each year.[1] Complications occur in up to 32%; major complications decrease quality of life and increase health care costs by as much as $11,000.[2,3] Accurate, personalized predictions of postoperative complications can inform shared decisions between patients and surgeons regarding prognosis, the appropriateness of surgery, prehabilitation strategies targeting modifiable risk factors (e.g., smoking cessation), and postoperative resource use (e.g., triage to intensive care or general wards), suggesting opportunities to augment clinical risk prediction with objective, machine learning-enabled decision-support.

Most existing perioperative predictive analytic decision-support tools are hindered by suboptimal performance, time constraints imposed by manual data entry requirements, and lack of intraoperative data and clinical workflow integration.[4-9] These challenges are theoretically mitigated by automated deep learning models that capture latent, nonlinear data structure and relationships among raw feature representations in large datasets,[10] now widely available in electronic health records (EHRs).[11] Despite these potential advantages,[12-19] deep learning using the full spectrum of preoperative and intraoperative, patient-specific EHR data to predict postoperative complications has not been previously reported. Recognition that deep learning models with high overall accuracy are nevertheless capable of egregious errors, along with their lack of interpretability, have invited skepticism regarding the clinical application of deep learning-enabled decision-support; model interpretability and uncertainty-awareness mechanisms have the potential to improve clinical applicability, but their efficacy remains unclear.

Using a longitudinal cohort of 56,242 patients who underwent 67,481 inpatient surgeries, we test the hypotheses that deep learning models would outperform random forest models in predicting postoperative complications using both preoperative and intraoperative physiological time series data. We also explore the utility of multi-task learning[20] by training a single deep learning model on several postoperative complications simultaneously to improve model efficiency, integrated gradients to promote model interpretability, and uncertainty metrics that represent variance across predictions.

## RESULTS

### *Participant Baseline Characteristics and Outcomes*

Cohort characteristics are summarized in Table 1 and detailed cohort statistics are presented in Supplementary Tables S1-S4. The overall study population had mean age 56 years and 50% were female. In the validation cohort of 20,293 surgical procedures, the incidence of complications was: 33.3% prolonged ICU stay (for 48 hours or more), 7.8% prolonged mechanical ventilation, 20.2% neurological complications, 16.9% acute kidney injury, 16.3% cardiovascular complications, 5.4% venous thromboembolism, 21.4% wound complications, 8.7% sepsis, and 1.6% in-hospital mortality. The distribution of complications was similar between development and validation cohorts.

## *Multi-Task Learning Improved Efficiency without Compromising Predictive Performance*

For deep learning models trained on preoperative data alone, there were no significant differences between multi-task outcome-specific models. For models trained on intraoperative time series alone, the multi-task model yielded significantly higher AUROC for sepsis (0.80 [95% confidence interval 0.78-0.81] vs. 0.78 [0.77-0.79]) and venous thromboembolism (0.74 [0.72-0.75] vs. 0.71 [0.69-0.73]). Using all available preoperative and intraoperative data, the multi-task postoperative model yielded somewhat higher AUROC for prolonged mechanical ventilation, sepsis, venous thromboembolism, and in-hospital mortality, and lower AUROC for prolonged ICU stay, wound complications, neurological complications, and acute kidney injury, though the differences were not statistically significant. A comprehensive AUROC comparison between individual models and multi-task learning is shown in Fig. 1a-c. Given that multi-task models had marginally stronger performance and have a reduced computational requirements and training times compared with nine individual models, the multi-task approach is used henceforth as our deep learning-based postoperative model, unless stated otherwise. Full results are shown in Supplementary Table S5.

## *Deep Learning Outperformed Random Forests*

Both deep learning and baseline models used the same feature sets with one exception: due to the nature of sequential deep learning methods, our deep intraoperative models processed the entire physiological time series minute-by-minute, whereas the baseline intraoperative and postoperative models required extraction of summary statistics. A full list of random forest time series features is described in Supplementary Table S6. A full comparison among all models, performance metrics, and complication outcomes is described in Supplementary Methods and Supplementary Table S5.

*Preoperative Models*

The deep multi-task model trained only on static, preoperative descriptors yielded higher AUROC compared with baseline random forest models for all nine outcomes, with significant performance increases for prolonged mechanical ventilation (0.90 [0.89-0.90] vs. 0.86 [0.85-0.87]), wound complications (0.77 [0.76-0.78] vs. 0.73 [0.72-0.73]), neurological complications (0.85 [0.84-0.85] vs. 0.83 [0.82-0.83]), cardiovascular complications (0.81 [0.80-0.81] vs. 0.78 [0.77-0.79]), acute kidney injury (0.82 [0.81-0.83] vs. 0.80 [0.79-0.80]), venous thromboembolism (0.82 [0.81-0.83] vs. 0.78 [0.76-0.79]), and in-hospital mortality (0.89 [0.88-0.91] vs. 0.84 [0.82-0.86]).

*Intraoperative Models*

Using intraoperative time series input data alone, multi-task deep learning yielded higher AUROC for all complications except prolonged ICU stay, for which AUROC was equivalent. Significant AUROC improvements were yielded for wound complications (0.62 [0.61-0.63] vs. 0.59 [0.58-0.60]), acute kidney injury (0.74 [0.73-0.74] vs. 0.71 [0.70-0.72]), venous thromboembolism (0.74 [0.72-0.75] vs. 0.67 [0.66-0.69]), and in-hospital mortality (0.88 [0.86-0.89] vs. 0.79 [0.77-0.82]).

*Postoperative Models*

The deep postoperative multi-task model trained on all available data yielded significant higher AUROC compared with a random forest baseline model for eight of nine complications, including prolonged ICU stay (0.91 [0.91-0.92] vs. 0.90 [0.89-0.90]), wound complications (0.77 [0.77-0.78] vs. 0.71 [0.70-0.72]), neurological complications (0.85 [0.85-0.86] vs. 0.81 [0.80-0.81]), cardiovascular complications (0.85 [0.85-0.86] vs. 0.82 [0.81-0.83]), sepsis (0.88 [0.87-0.88] vs. 0.85 [0.84-0.85]), acute kidney injury (0.82 [0.82-0.83] vs. 0.80 [0.79-0.81]), venous thromboembolism (0.83 [0.81-0.84] vs. 0.75 [0.74-077]), and in-hospital mortality (0.92 [0.91-0.93] vs. 0.83 [0.81-0.86]). The deep multi-task model yielded somewhat higher AUROC for prolonged mechanical ventilation, but the difference was not statistically significant. A full AUROC comparison between deep learning models and random forest baselines is shown in Fig. 1a-c and Supplementary Table S5.

***Deep Postoperative Models Outperformed Deep Preoperative Models***

Compared with deep preoperative models, deep postoperative models had significantly higher AUROC for prolonged ICU stay (0.91 [95% confidence interval 0.91-0.92] vs. 0.88 [0.88-0.89]), prolonged mechanical ventilation (0.93 [0.92-0.94] vs. 0.90 [0.89-0.90]), and cardiovascular complications (0.85 [0.85-0.86] vs. 0.81 [0.80-0.81]). A full comparison is shown in Fig. 1d. Using deep multi-task preoperative predictions as a benchmark, the deep multi-task postoperative models made significant overall reclassification improvements for prolonged ICU stay (overall, correctly reclassified 3.7% of all surgical encounters, p<0.01), prolonged mechanical ventilation (overall, correctly reclassified 4.8%, p<0.01), and cardiovascular complications (overall, correctly reclassified 0.3%, p<0.01). There were no statistically significant declines in reclassification. In some cases, deep models for individual complications yielded better net reclassification indices than multi-task models, including wound complications (-1.7% vs. -2.9%, p<0.01) and cardiovascular complications (2.8% vs. 0.3%, p<0.01). Full net reclassification results are shown in Supplementary Table S7. Detailed statistics for absolute and relative risk between preoperative and postoperative models are shown in Supplementary Table S8, and analyses of risk group transitions are shown in Supplementary Tables S9 and S10.

***Model Uncertainty***

We applied the method of Monte Carlo dropout to derive measures of prediction uncertainty, representing variance across predictions, for each of our deep learning models. Uncertainty results for each prediction phase and training procedure are shown in Table 2, where uncertainty is expressed as prediction variance over 100 stochastic trials using dropout at inference time. Interestingly, models trained only using intraoperative data resulted in the lowest uncertainty for each postoperative complication. Within each outcome and prediction phase, individual models yielded lower predictive uncertainty compared with multi-task model counterparts. Using the models with the least uncertain training scheme for each outcome and prediction phase, postoperative predictions were less uncertain than preoperative predictions for prolonged mechanical ventilation, wound complications, cardiovascular complications, and in-hospital mortality; postoperative uncertainty was higher for the remaining five complications.

### *Model Interpretability*

We applied integrated gradients to our multi-task deep learning postoperative prediction model. The top 10 features per complication outcome for every sample in the validation cohort are shown with corresponding attribution scores in Table 3. Importance distribution among the top 10 features per complication are visualized in Supplementary Fig. S1, and distributions of feature importance values grouped by input and feature type are visualized in Supplementary Fig. S2 and S3. The important feature lists, as described in subsequent sections, are consistent with medical knowledge, experience, and evidence, establishing an important element in gaining the trust of patients and clinicians.[21]

#### *Prolonged ICU* Stay

The most important feature was peak inspiratory pressure; the presence of such a value indicates the performance of mechanical ventilation, and higher values could indicate intrinsic lung disease, proximal airway or breathing tube narrowing or obstruction, or the transmission of increased intra-abdominal pressure, each of which suggest greater illness severity. The next two most important features were heart rate and blood oxygen saturation, both of which are major determinants of cardiac output and oxygen delivery.

*Prolonged Mechanical Ventilation*

Peak inspiratory pressure and heart rate were again top features, along with fraction of inspired oxygen, the number one feature. This result is consistent with prior observations that most etiologies of hypoxemia improve with increasing fraction of inspired oxygen, apart from right-to-left shunt, which is often accompanied by another pathophysiologic process that is responsive to higher fraction of inspired oxygen. Temporal feature attributions for physiological intraoperative time series from an example patient requiring prolonged mechanical ventilation are shown in Fig. 2, with another example for a patient developing cardiovascular complications shown in Supplementary Fig. S4.

*Wound Complications*

The major factors affecting wound complications (i.e., infection, dehiscence, and non-healing) are the type of surgery its associated degree of wound contamination.[22,23] These factors are aligned with the top four important features for wound complication prediction: primary procedure, surgeon specialty, attending surgeon, surgery type, and scheduled surgery room. Although body mass index is unexpectedly missing from the top 10 feature list, several other factors relate to known risk factors for wound complications, including malnutrition, long duration of surgery, blood loss, and anemia.

*Neurological Complications*

Similar to wound complications, neurological complications are primarily a function of type of surgery; neurosurgical procedures typically involve pre-existing neurological pathology and confer above-average risk for postoperative neurological pathology relative to other types of surgery. Accordingly, primary procedure and surgery type were the top two important features in predicting neurological complications.

*Cardiovascular Complications*

Cardiovascular complications may be caused by or lead to cardiac and respiratory pathophysiology, primarily measured by cardiac and respiratory vital signs and mechanical ventilator measurements.[24] Consistent with these phenomena, the top five important features for cardiovascular complications were systolic blood pressure, peak inspiratory pressure, blood oxygen saturation, heart rate, and diastolic blood pressure.

*Sepsis*

Important features for sepsis were similar to those of wound complications, with the exception of heart rate, which was the most important feature for sepsis. One might expect that fever, leukocytosis, and hypotension would be important features in predicting sepsis, but it is possible that these elements would occur later after surgery when sepsis was developing as a postoperative complication, and they can also represent sterile postoperative inflammation from tissue damage without infection. Heart rate variability, which would be learned from intraoperative time series heart rate values, is well established as a strong predictor of sepsis and associated adverse outcomes.[25,26]

*Acute Kidney Injury*

Serum creatinine is the primary method for measuring kidney function among hospitalized patients and tends to be more reliable than volume of urine output, which is difficult to record accurately in the absence of an indwelling bladder catheter. Accordingly, the number one important feature in predicting acute kidney injury was serum creatinine. Several other important features represented kidney perfusion or red blood cell production, which is affected by the endogenous renal hormone erythropoietin.

*Venous Thromboembolism*

Major risk factors for venous thromboembolism are encompassed by Virchow's triad of vessel injury, altered blood flow, and hypercoagulability.[27] These elements are represented in two of the top three important features for predicting venous thromboembolism (i.e., primary procedure and serum prothrombin time), as well as several other variables in the top 10 feature list.

## DISCUSSION

In predicting postoperative complications among adult patients undergoing major, inpatient surgery, deep neural networks outperformed random forest classifiers, exhibiting strongest performance when leveraging the full spectrum of preoperative and intraoperative EHR data. Intraoperative physiological time-series had meaningful associations with postoperative patient outcomes, suggesting that prediction models augmented with intraoperative data may have utility for routine clinical tasks such as sharing prognostic information with patients and caregivers and making clinical management decisions regarding triage destination and resource use after surgery. Deep models maintained high performance using efficient multi-task methods predicting nine complications simultaneously, rather than predicting individual complications with separate models that require extra training time. Uncertainty metrics revealed that variance across model predictions is lowest when using intraoperative data alone, consistent with the perspective that many preoperative EHR predictor variables represent clinician decision-making (e.g., the lack of preoperative bilirubin values indicates a decision to forego hepatic function testing) rather than pure physiology, and therefore introduce greater variance in predictions. Finally, applying integrated gradients interpretability methods elucidated feature importance patterns that were biologically plausible and consistent with medical knowledge, experience, and evidence, harboring the potential to gain trust from patients and clinicians.[21]

Previous studies have established that for many clinical prediction tasks, deep neural networks outperform other methods, such as logistic regression classifiers.[28,29] Parametric regression equations often fail to accurately represent complex, non-linear associations among input variables, limiting their predictive performance. More than thirty years ago, Schwartz et al.[30] suggested that human disease is too broad and complex to be accurately represented by rule-based algorithms, and that machine learning models obviate this limitation by learning from data. In our study, deep learning also outperformed random forest models, likely because the deep models capitalized on the availability of intraoperative time series data. As EHR data volumes expand, deep learning healthcare applications gain greater potential for clinical application.[31]

However, this will require integration with real-time clinical workflow. Therefore, it seems prudent to design models that make updated predictions as EHR data become available. We sought to achieve this objective by using recurrent neural networks that can update their predictions when new data becomes available. Our results suggest that these models would perform well in prospective clinical settings.

Multi-task methods did not yield predictive performance advantages in our study, but it has yielded performance advantages in prevoius studies. Muti-task learning can improve model generalizability by penalizing the exploration of certain regions of the available function space, thus reducing overfitting from the false assumption that data noise is sparse or absent. This has been demostrated by Si and Roberts[32] in applying CNN multi-task learning to word embeddings in MIMIC-III clinical notes data, demonstrating that multi-task learning models outperformed single-task models in predicting mortality within 1, 3, 5, and 20 different timeframes. In addition, multi-task learning can act as a regulizer for learning classifiers from a finite set of examples by penalizing complexity in a loss function, as demonstrated by Harutyunyan et al.[20] in predicting mortality and physiological decompensation among ICU patients in the publicly available MIMIC-III database.[33] However, multi-task learning was not advantageous for phenotyping acute care conditions; the authors postulated that this occurred because phenotyping is multi-task by nature, i.e., already benefits from regularization across phenotypes. This may not hold true for rare, complex phenotypes, for which multi-task learning can reduce neural network sensitivity to hyperparameter settings (i.e., parameters that are set before learning begins), as demonstrated by Ding et al.[34] Properly applied, multi-task learning can improve model generalizability and classification in deep learning clinical prediction models, optimizing performance and usability across diverse settings and datasets, with the added advantage of reduced model training times relative to training multiple individual models.

One barrier to clinical adoption of deep learning clinical prediction models is difficulty interpreting outputs. Patients, caregivers, and clinicians may be more willing to incorporate model predictions in shared decision-making processes if they understand how and why a prediction was made and believe that the prediction is consistent with medical knowledge and evidence. Integrated gradients techniques attempt to explain predictions made by deep learning models, usually by feeding perturbed inputs to the model, evaluating effects on outputs, and using this information to quantify and convey feature importance. Sayres et al.[35] used integrated gradients to identify retinal image regions contributing to deep learning-based diabetic retinopathy diagnoses, which was associated with improved ophthalmologist diagnostic accuracy and confidence. These methods have the potential to facilitate clinical adoption of deep learning prediction models by allowing patients, caregivers, and clinicians to undertand how and why and output was produced. Finally, demonstrating low variance across predictions with uncertainty metrics could assuage well-founded patient and clinician fears that an individual model output represents a rare but egregious prediction error, for which deep learning models are infamous.

This study was limited by its single-institution, retrospective design. Although multi-task functions may reduce overfitting, the use of data from a single institution limits generalizability. Our models have not been tested using prospective, real-time data, which may present data pre-processing challenges. Future research should seek prospective, multi-center validation of these findings. This will be difficult to perform until cloud sharing of standardized EHR data or federated learning are achieved at scale.[36] Finally, it remains unknown how the predictions generated by models presented herein would affect shared decision-making processes and patient outcomes.

In summary, deep learning yielded greater discrimination than random forests for predicting complications after major, inpatient surgery. Uncertainty metrics and predictive performance were optimal when leveraging the full spectrum of preoperative and intraoperative physiologic time-series data as predictor variables in an efficient multi-task deep learning model. Uncertainty-aware deep learning may have utility for understanding the probability that a prediction deviates substantially from usual predictions and represents a rare, major prediction error. Integrated gradients interpretability mechanisms identified biologically plausible important features. The accurate, interpretable, uncertainty-aware predictions presented herein require further investigation regarding their potential to augment surgical decision-making during preoperative and immediate postoperative phases of care.

## METHODS

All analyses were performed on a retrospective, single-center, longitudinal cohort of surgical patients that included data from both preoperative and intraoperative phases of care. We used deep learning and random forest models to predict the onset of nine major postoperative complications following surgery with three primary objectives: (1) compare deep learning techniques with random forest models in predicting postoperative complications, (2) compare deep learning predictions made at two phases of perioperative care: immediately before surgery (using preoperative data alone, referred to henceforth as preoperative prediction), and immediately after surgery by two different methods: (a) using intraoperative data alone (referred to henceforth as intraoperative prediction), and (b) using both preoperative and intraoperative data (referred to henceforth as postoperative prediction), and (3) explore the potential benefits of three novel deep learning techniques: (a) multi-task learning by training a single deep learning model on several postoperative complications compared with training separate models for each individual complication, (b) model interpretability with integrated gradients, and (c) model uncertainty-awareness by calculating variance across predictions.

The University of Florida Institutional Review Board and Privacy Office approved this study as an exempt study with waiver of informed consent (IRB # 201600223). Recommendations were followed from both Transparent Reporting of a multivariable prediction model for Individual Prognosis Or Diagnosis (TRIPOD[37]) guidelines and from best practices for prediction modeling from Leisman, et al.[38] All methods were performed in accordance with relevant guidelines and regulations.

***Data Source***

The University of Florida Integrated Data Repository was used as an honest broker to build a longitudinal dataset representing patients admitted to University of Florida Health between June 1st, 2014 and September 20th, 2020 who were at least 18 years of age and underwent at least one surgical procedure during hospitalization. The dataset was constructed by integrating electronic health records with other clinical, administrative, and public databases.[9] The resulting dataset included information on patient demographics, laboratory values, vital signs, diagnoses, medications, blood product administration, procedures, and clinical outcomes, as well as detailed intraoperative physiologic and monitoring data.

### *Predictors*

Our final cohort included electronic health record data from both before and during surgery. Preoperative models were trained on data available between one year prior to surgery and the day of surgery, prior to surgery start time (i.e., preoperative features alone). Intraoperative models were trained on data created during the surgical procedure (i.e., intraoperative features alone). Postoperative models were trained on data available between one year prior to surgery through the end of the surgical procedure (i.e., both preoperative and intraoperative features).

We identified 402 preoperative features, including demographic and socioeconomic indicators, planned procedure and provider information, Charlson comorbidities, and summary statistics of select medications, laboratory tests, and physiological measurements (e.g., vital signs such as heart rate and blood pressure) taken prior to a surgical procedure over one-year and one-week time windows. We calculated Charlson comorbidity indices using International Classification of Diseases (ICD) codes.[39] We modeled procedure types on ICD-9-CM codes with a forest structure in which nodes represent groups of procedures, roots represent the most general groups of procedures, and leaf nodes represent specific procedures. Medications were derived from RxNorm codes grouped into drug classes as previously described.

Intraoperative data consisted of 14 physiological measurements taken during surgery: systolic blood pressure, diastolic blood pressure, mean arterial pressure, heart rate, blood oxygen saturation (SpO2), fraction of inspired oxygen (FiO2), end-tidal carbon dioxide (EtCO2), tidal volume, respiration rate, peak inspiratory pressure (PIP), minimum alveolar concentration (MAC), temperature, urine output, and operative blood loss. These variables were presented to deep learning models as variable-length multivariate time series. For random forest models, a set of 49 statistical features were extracted from each encounter's intraoperative measurements. Supplementary Table S6 summarizes all input features and relevant preprocessing procedures.

***Participants***

We excluded patients with intraoperative mortality or who were missing the variables necessary to classify postoperative complications. If a single patient's hospital encounter included more than one surgery, only the first surgery during that encounter was included in our analyses. Our final dataset included 56,242 patients who underwent 67,481 surgeries. Supplementary Fig. S5 illustrates derivation of the study population and cohort selection criteria.

***Outcomes***

We used several different machine learning methods to model the risk of nine postoperative complications: prolonged intensive care unit stay (greater than 48 hours), prolonged mechanical ventilation requirement (greater than 48 hours), neurological complications, cardiovascular complications, acute kidney injury, sepsis, venous thromboembolism, wound complications, and in-hospital mortality.

### *Sample Size*

We chronologically divided our perioperative cohort into a development set of 47,188 surgeries occurring between June 1st, 2014 through November 26th, 2018, and a validation set of 20,293 surgeries occurring between November 27th, 2018 through September 20th, 2020. All models were trained on the development set; all results were reported for the validation set. For deep learning models, we used 10% of the development set for early stopping.

Using a validation cohort of 20,293 surgeries, the overall sample size allows for a maximum width of the 95% confidence interval for area under the receiver operating characteristic curve (AUROC) to be between 0.01 to 0.03 for postoperative complications with prevalence ranging between 5.4% and 33.3% for AUROC of 0.80 or higher. The sample size allows for a maximum width of 0.06 for hospital mortality given 1.6% prevalence.

### *Predictive Analytic Workflow*

The postoperative models update preoperative risk predictions using data collected during surgery. This workflow emulates clinical scenarios in which patients' preoperative information is enriched by the influx of new data from the operating room. The model consists of two main preoperative and intraoperative layers, each containing a data transformer core and a data analytics core.[9] The data transformer integrates data from multiple sources, including EHR data with zip code links to US Census data for patient neighborhood characteristics and distance from the hospital. The data transformer then performs preprocessing and feature transformation steps to optimize the data for analysis.

The 402 preoperative features contained 341 continuous features, 42 binary features, and 19 nominal features. Of the 19 nominal features, 13 contained fewer than 5 levels and were one-hot encoded as zero vectors of dimension equal to number of levels, with level indicators equal to one. The remaining six nominal features (ZIP code, attending surgeon, primary procedure, scheduled operating room, surgery type, and surgeon specialty) were represented as unique integer identifiers ranging from zero to the number of levels minus one. Implicit variable representations were learned as part of the model training process. Continuous preoperative feature observations that fell below the $1^{st}$ or above the $99^{th}$ percentiles were capped to the $1^{st}$ and $99^{th}$ percentile values, respectively. Temporal preoperative features denoting the day and month of admission were transformed into two individual continuous features each through the use of sinusoidal functions based on the respective frequency of days or months, which encoded relative differences between time points (e.g., Sunday is close to Monday, and December is close to January).

Intraoperative measurements were identified as those falling between anesthesia start and stop times for a given procedure. Fixed-interval multivariate physiological time series were constructed for each procedure by resampling measured values to a frequency of one minute, which represented the highest recorded frequency across all intraoperative features. For a given surgical procedure which had at least one measurement of a given feature, any gaps in that feature's time series were imputed via linear interpolation in both directions. As surgeries vary in duration, each sample included a multivariate time series of length $T$ minutes. Blood loss sum, urine output sum, and duration of surgery were included as static postoperative features.

Missing continuous features were imputed with the median of each feature value in the development cohort. For static preoperative descriptors, this represented a single number; for intraoperative time series, this was only performed when a single feature value did not exist, and the median value was imputed at every one-minute time step for the full duration of surgery. Missing preoperative nominal features were replaced with a distinct "missing" category.

To preserve patterns of missingness which may be informative,[40] for each sample we derived a preoperative binary presence mask over all continuous and binary input variables that indicated whether a given value was observed or imputed. These missingness indicators were concatenated with their respective original measurements. For a given cohort set of size $N$ encounters, initial continuous and binary preoperative features were represented as a matrix of descriptors $P^{Nx383}$. With a missingness mask of size $P_{mask}^{Nx383}$, concatenation resulted in a final continuous and binary preoperative feature set of 766 numerical preoperative descriptors for each sample. Nominal preoperative features did not require a missingness mask, as missing values were transformed into a distinct categorical level. The 13 nominal variables that were one-hot encoded were concatenated with the above numerical preoperative representation, and the 6 nominal features with greater than 5 levels were internally embedded by the model. Multivariate time series missingness masks were computed and concatenated at each one-minute intraoperative timestep; for a single surgical time series $x^{Tx12}$ of length $T$ including our 12 temporal physiological measurements, the concatenation of these per-timestep masks resulted in a final input time series $x^{Tx24}$ of 24 intraoperative predictors at each timestep. All continuous input variables, both preoperative and intraoperative, were z-normalized to zero mean and unit variance based on values from the development set.

Following these processing steps, each surgical encounter was represented by four distinct sets of variables: a set of numerical preoperative features, a set of nominal preoperative features to be internally embedded by the model, a multivariate time series of length $T$ composed of physiological measurements, and a set of static surgical features collected at the end of surgery. The length of intraoperative time series varied depending on surgery duration, and our deep learning models were designed to process the full scope of intraoperative physiological measurements.

In the data analytics core, deep learning and random forest models were trained to predict nine postoperative complications following a surgical procedure. Clinically, predictions made by preoperative models can inform patients, caregivers, and surgeons regarding risks of undergoing surgery, and estimate the utility of risk reduction strategies for specific complications (e.g., preoperative smoking cessation, perioperative renal protection bundles, and wound closure techniques). Intraoperative events can influence risk for complications (e.g., operative blood loss requiring allogenic blood transfusion increases risk for septic complications and intraoperative hypotension increases risk for acute kidney injury). Therefore, we generated intraoperative models to predict complications using data obtained during surgery. At the end of surgery, clinicians must reassess the patient's prognosis, convey this information to the patient and their caregivers, and make clinical management decisions accordingly (e.g., a patient at high risk for cardiovascular complications may benefit from postoperative admission to an intensive care unit or continuous cardiac telemetry on a general hospital ward). At the end of surgery, it seems prudent to consider both baseline preoperative risk as well as the potential influence of intraoperative events to make updated predictions of postoperative complications. This is accomplished by our postoperative models.

As a technical explanation of deep learning fundamentals is beyond the scope of this study, we refer interested readers to the comprehensive work by Goodfellow et al.[41] Our final postoperative deep learning model can be conceptualized as a composition of two sub-models: one for processing preoperative features, and one for processing intraoperative features. Reported preoperative results (i.e., predicting postoperative complications using preoperative features alone) were obtained by only using the data representation from the preoperative sub-model; likewise, reported intraoperative results were obtained by only using the data representation from the intraoperative sub-model. The postoperative model used a transformed concatenation of both preoperative and intraoperative data representations (Supplementary Fig. S6).

The preoperative sub-model was composed of a dual pipeline for processing and representing numerical features and nominal features with greater than 5 levels. A representation of all six index-encoded nominal input features was obtained by concatenating individual nominal feature representations, each of which were the result of a learned, multidimensional per-feature embedding lookup table, and passing the concatenated result through a fully connected layer. A representation of all numerical preoperative variables was obtained by passing the input features through a fully connected layer. A complete preoperative encounter representation was obtained by concatenating both continuous and nominal input feature representations and passing the result through a final fully connected layer.

In the multi-task setting, this preoperative data representation was passed through nine branches corresponding to our nine postoperative complication outcomes. Each branch contained one outcome-specific fully connected layers followed by a sigmoid activation function to produce a per-outcome prediction score, interpreted as the probability of a preoperative patient developing a given postoperative complication.

The primary driving force behind the intraoperative sub-model was a bidirectional recurrent neural network (RNN) with gated recurrent units (GRU). A patient's intraoperative time series was passed through the RNN twice, once in chronological order and once in reverse order. Time step representations were generated by concatenating the RNN hidden states from the forward and backward passes. An attention mechanism was applied to the bidirectional sequence representations. Briefly, an attention mechanism for classification allows a model to assign importance scores to individual timesteps of a representation sequence such that the importance-weighted sequence is summed into a single context vector that is an optimal representation for a given predictive task. Attention allows a model to learn to focus exclusively on timesteps that are important for classification decisions. In our multi-task model, we implemented a separate attention mechanism for each of the nine postoperative complications. Using a shared representation of an intraoperative sequence from the RNN, each attention component formulated a separate perspective of the sequence aligned with each outcome of interest.

Our complete deep learning model, which we refer to as the postoperative model, includes both the preoperative and intraoperative sub-models described in this section. The postoperative model is trained end-to-end and consists of concatenating both the static preoperative representation (the output of the preoperative sub-model) with the outcome-specific intraoperative representation (the output of the intraoperative sub-model for a given outcome) and passing this combined feature representation through the same set of nine classification branches as the sub-models.

In our experiments and reported results, we use a nominal preoperative variable embedding size of 64, fully connected layers size of 64 (except for final task output layers, which have size 1), hidden dimension of 64, Adam optimizer with learning rate of 0.001, L2 regularization of 0.01, batch size of 64, RELU activation, and patience of 4 used for early stopping based on the validation data set.

To determine whether the deep learning models offered a performance advantage over traditional predictive analytic methods, we assessed the performance of a baseline random forest classifier using the same preoperative and intraoperative input feature sets as the deep learning models, with predictions made at the same time points. Nominal preoperative features, which were index-encoded before passing through the deep model, were instead one-hot encoded before feeding into the baseline random forest model. Intraoperative time series were fed to the baseline model by way of 49 summary statistics, capturing static attributes and patterns of variability for each variable. These features are described in Supplementary Table S6.

To account for class imbalance among the nine postoperative complication outcomes, both deep learning and baseline models were trained using outcome-specific class weights that were inversely proportional to their respective frequencies in the training set. Functionally, this ensures greater model focus on minority class samples.

The postoperative complication predictions from all deep neural networks trained under each surgical phase (preoperative, intraoperative, postoperative) and training scheme (individual models, multi-task learning) were analyzed with Monte Carlo dropout, approximating Bayesian inference and providing a quantitative measure of uncertainty for neural network predictions.[42] By enabling randomized dropout during model inference and aggregating resulting predictions over several experimental trials, a pseudo model ensemble is generated with partially randomized neural network connections. In our experiments, we perform 100 trials with stochastic dropout applied during inference and compute the mean complication risk and resulting prediction variance as a measure of model uncertainty.

We apply the method of integrated gradients to our final postoperative multi-task model to illuminate specific input features that yielded the largest impacts on predicting each of the nine complication outcomes. A complete discussion of this technique is beyond the scope of this study; we refer interested readers to the work of Sundarajan et al.[43] Briefly, integrated gradients is a comparative technique for local interpretability, centered around the analysis of model outputs based on a given input and corresponding baseline values, and assigns attributions values to every input feature. In theory, features most influential to a given prediction will receive larger attribution values, and taken over an entire population, this can reveal the importance of certain features which drive the model predictions. We use a zero-vector reference value for such computations, and as all variables are Z-normalized to zero mean and unit variance; such a reference can be viewed as the per-variable mean value across the entire cohort.

***Model Validation***

All models were trained on the development set of 47,188 surgeries occurring between June 1st, 2014 through November 26th, 2018. Models were evaluated on the validation set of 20,293 surgical procedures occurring between November 27th, 2018 through September 20th, 2020. For each model performance metric, ninety-five percent nonparametric confidence intervals were calculated using 1,000 bootstrapped samples with replacement.

***Model Performance***

Model performance was evaluated by sensitivity, specificity, positive predictive value (PPV), negative predictive value (NPV), accuracy, area under the precision-recall curve (AUPRC), and area under the receiver operating characteristic curve (AUROC). Reported metrics include class predictions based on Youden's index threshold on predicted risk scores, which maximizes sensitivity and specificity, as the cutoff point for low versus high risk.[44]

When predicting rare events, models can exhibit deceivingly high accuracy by predicting negative outcomes in predominantly negative datasets.[45] False negative predictions of postoperative complications may be especially detrimental because patients, caregivers, and surgeons could unknowingly agree to perform prohibitively high-risk surgery, miss opportunities to mitigate preventable harm through prehabilitation and other risk-reduction strategies, and under-triage high-risk patients to general hospital wards with infrequent monitoring, when close monitoring in an intensive care unit would be safer. Therefore, model performance was evaluated by calculating area under the precision-recall curve (AUPRC), which is adept at evaluating the performance of models predicting rare events.[46] In addition, Net Reclassification Improvement (NRI) indices were used to describe and quantify correct and incorrect reclassifications by deep learning models.[47] For all performance metrics, we used bootstrap sampling and non-parametric methods to obtain 95% confidence intervals.

## ACKNOWLEDGEMENTS

T.J.L. was supported by the National Institute of General Medical Sciences of the National Institutes of Health under Award Number K23 GM140268. A.B. was supported by R01GM110240 from the National Institute of General Medical Sciences (NIH/NIGMS), R01EB029699 and R21EB027344 from the National Institute of Biomedical Imaging and Bioengineering (NIH/NIBIB), R01NS120924 from the National Institute of Neurological Disorders and Stroke (NIH/NINDS), and by R01DK121730 from the National Institute of Diabetes and Digestive and Kidney Diseases (NIH/NIDDK). T.O.B. was supported by K01DK120784, R01DK123078, and R01DK121730 from the National Institute of Diabetes and Digestive and Kidney Diseases (NIH/NIDDK), R01GM110240 from the National Institute of General Medical Sciences (NIH/NIGMS), R01EB029699 from the National Institute of Biomedical Imaging and Bioengineering (NIH/NIBIB), and R01NS120924 from the National Institute of Neurological Disorders and Stroke (NIH/NINDS). P.R. was supported by National Science Foundation CAREER award 1750192, R01EB029699 and R21EB027344 from the National Institute of Biomedical Imaging and Bioengineering (NIH/NIBIB), R01GM110240 from the National Institute of General Medical Science (NIH/NIGMS), R01NS120924 from the National Institute of Neurological Disorders and Stroke (NIH/NINDS), and by R01DK121730 from the National Institute of Diabetes and Digestive and Kidney Diseases (NIH/NIDDK). The content is solely the responsibility of the authors and does not necessarily represent the official views of the National Institutes of Health.

## AUTHOR CONTRIBUTIONS

B.S., T.L., M.R., T.O.B., and A.B. had full access to the data in the study and take responsibility for the integrity of the data and accuracy of the data analysis. Study design performed by B.S., T.L., A.B. and P.R. Manuscript was drafted by B.S. and T.L. Data extraction and processing was performed by B.S. and M.R. Experiments were conducted by B.S. and clinical interpretation was performed by T.L. and G.U. Funding was obtained by A.B. and P.R. Administrative, technical, material support was provided by A.B. and P.R. Study supervision was performed by A.B. and P.R. All authors contributed to the acquisition, analysis, and interpretation of data. All authors contributed to critical revision of the manuscript for important intellectual content.

## DATA AVAILABILITY

Data is available from the University of Florida Institutional Data Access/Ethics Committee for researchers who meet the criteria for access to confidential data and may require additional IRB approval.

## COMPETING INTERESTS

The authors declare no competing interests.

## FIGURES

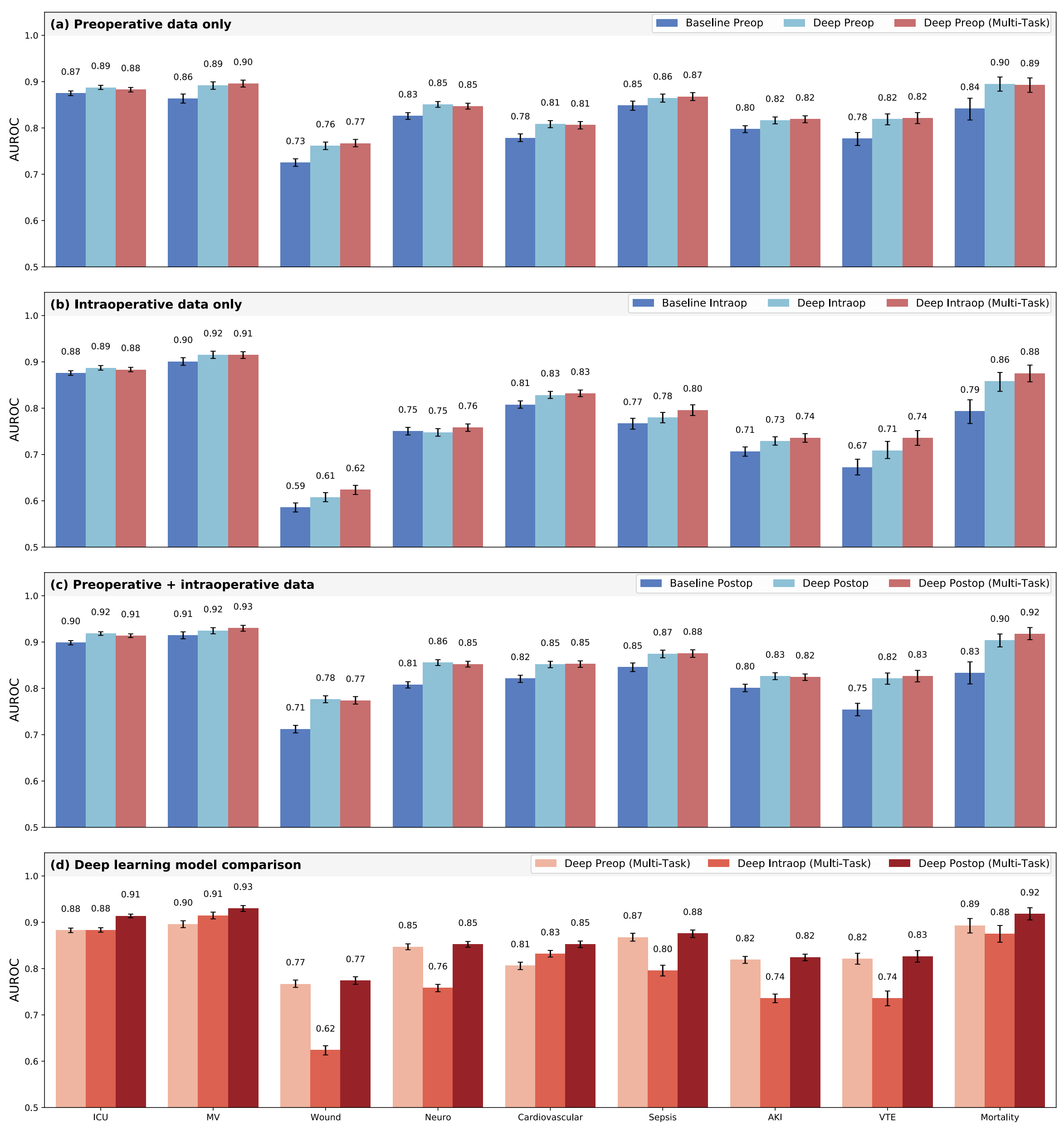

**Figure 1. Classification accuracy compared with baseline models.** Shown are area under the receiver operating characteristic curve (AUROC) results for random forest models, individual deep learning models independently trained on each outcome, and a combined multi-task jointly trained on all outcomes, using only preoperative features (a), only intraoperative features (b), and both preoperative and intraoperative features (c). A comparison of multi-task deep learning results at three stages of prediction is shown in

(d).

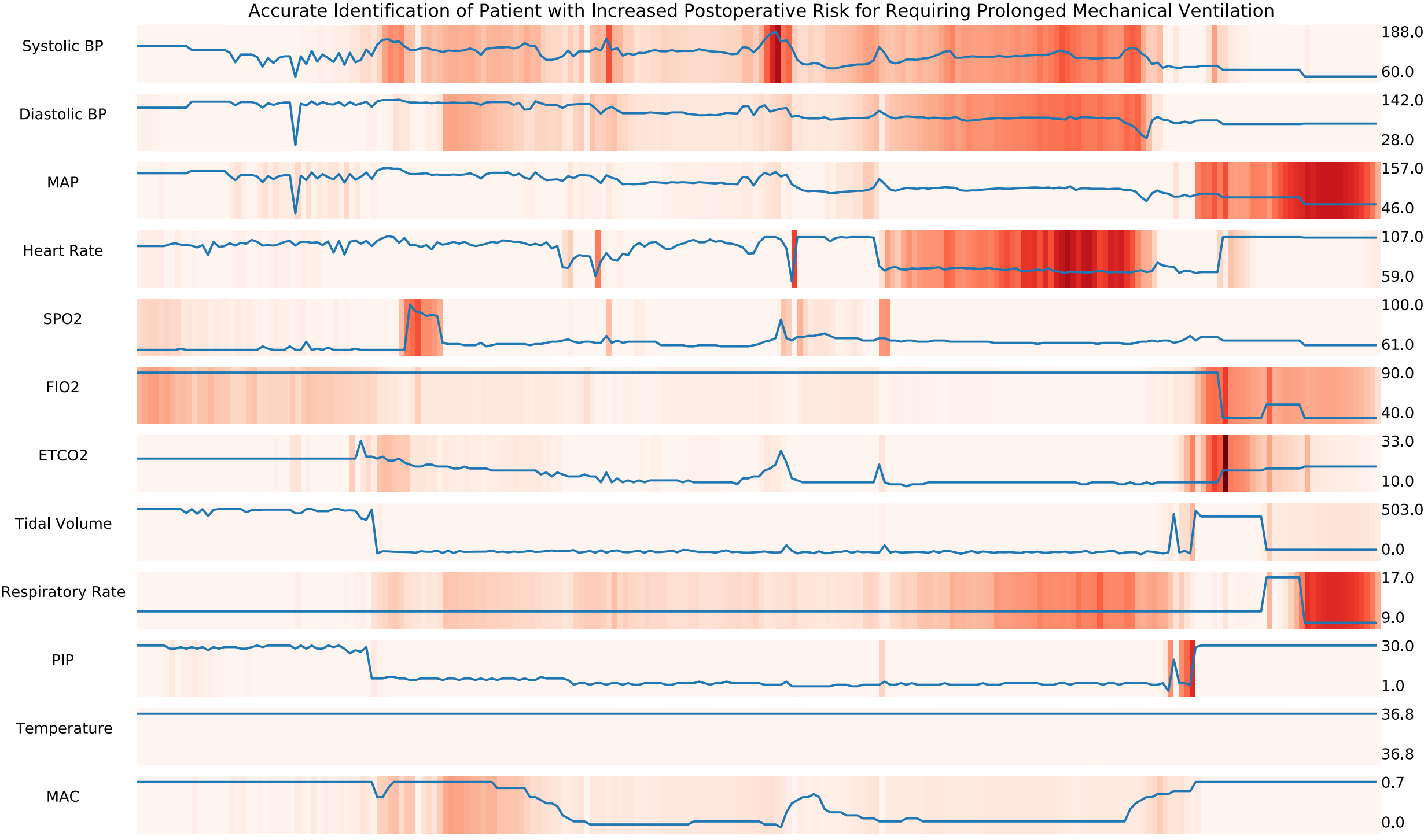

**Figure 2. Temporal integrated gradients feature attributions for example patient experiencing prolonged mechanical ventilation.** The multi-task deep learning model correctly predicted elevated risk of prolonged mechanical ventilation after integrating multivariate intraoperative time series. Physiological time series labeled by variable (left) and value range (right). Implementation of integrated gradients highlighted physiological patterns important for updated risk prediction, including a rapid increase in heart rate and ETCO2, fluctuations in PIP, and changes in SPO2. ETCO2: end-tidal carbon dioxide; PIP: peak inspiratory pressure; SPO2: blood oxygen saturation.

## TABLES

### Table 1. Summary of development and validation cohorts.

| | **Development Cohort** (6/1/2014 - 11/26/2018) | **Validation Cohort** (11/27/2018 - 9/20/2020) |
|---|---|---|
| **Patients, n** | 38621 | 17621 |
| **Hospital encounters, n** | 47188 | 20293 |
| **Age, years, median (25th, 75th)** | 59.0 (45.0, 69.0) | 61.0 (47.0, 71.0) |
| **Length of stay, days, median (25th, 75th)** | 4.1 (2.2, 7.9) | 4.3 (2.2, 8.4) |
| **Length of surgery, hours, median (25th, 75th)** | 3.1 (2.2, 4.6) | 3.2 (2.3, 4.7) |
| **Emergent Admission, n (%)** | 16706 (35.4%) | 7491 (36.9%) |
| **Charlson comorbidity index, median (25th, 75th)** | 4.0 (2.0, 6.0) | 4.0 (2.0, 6.0) |
| **Sex, n (%)** | | |
| Female | 23716 (50.3%) | 10005 (49.3%) |
| Male | 23472 (49.7%) | 10288 (50.7%) |
| **Race, n (%)** | | |
| White | 37047 (78.5%) | 15942 (78.6%) |
| African American | 6562 (13.9%) | 2759 (13.6%) |
| Other/Unknown | 3579 (7.6%) | 1592 (7.8%) |
| **Admission Type, n (%)** | | |
| Medicine | 20893 (44.3%) | 8277 (40.8%) |
| Surgery | 17899 (37.9%) | 7806 (38.5%) |
| Other | 8396 (17.8%) | 4210 (20.7%) |
| **Postoperative Complications, n (%)** | | |
| Prolonged ICU Stay (> 2 Days) | 12980 (27.5%) | 6765 (33.3%) |
| Prolonged Mechanical Ventilation (> 2 Days) | 3512 (7.4%) | 1574 (7.8%) |
| Wound Complications | 6782 (14.4%) | 4347 (21.4%) |
| Neurological Complications | 7273 (15.4%) | 4107 (20.2%) |
| Cardiovascular Complications | 5655 (12.0%) | 3301 (16.3%) |
| Sepsis | 3445 (7.3%) | 1775 (8.7%) |
| Acute Kidney Injury | 6894 (14.6%) | 3438 (16.9%) |
| Venous Thromboembolism | 2008 (4.3%) | 1101 (5.4%) |
| In-Hospital Mortality | 788 (1.7%) | 321 (1.6%) |

## Table 2. Deep model uncertainty metrics aggregated over 100 Monte Carlo dropout iterations.

| Outcome | Prediction Point | Model Type | Mean Uncertainty (Variance x $10^3$) | Mean AUROC |
|---|---|---|---|---|
| Prolonged ICU Stay | Preop | Individual | 3.471 | 0.887 |
| | | Multi-Task | 5.907 | 0.883 |
| | Intraop | Individual | 1.178 | 0.887 |
| | | Multi-Task | 1.401 | 0.884 |
| | Postop | Individual | 3.675 | 0.919 |
| | | Multi-Task | 4.785 | 0.914 |
| Prolonged MV | Preop | Individual | 2.695 | 0.892 |
| | | Multi-Task | 4.040 | 0.896 |
| | Intraop | Individual | 0.851 | 0.915 |
| | | Multi-Task | 1.274 | 0.916 |
| | Postop | Individual | 2.345 | 0.925 |
| | | Multi-Task | 3.122 | 0.931 |
| Wound | Preop | Individual | 3.058 | 0.761 |
| | | Multi-Task | 6.191 | 0.767 |
| | Intraop | Individual | 0.553 | 0.608 |
| | | Multi-Task | 1.077 | 0.624 |
| | Postop | Individual | 2.884 | 0.777 |
| | | Multi-Task | 5.363 | 0.774 |
| Neuro | Preop | Individual | 2.194 | 0.851 |
| | | Multi-Task | 4.856 | 0.847 |
| | Intraop | Individual | 1.553 | 0.748 |
| | | Multi-Task | 1.326 | 0.758 |
| | Postop | Individual | 2.723 | 0.855 |
| | | Multi-Task | 4.270 | 0.852 |
| Cardiovascular | Preop | Individual | 2.331 | 0.809 |
| | | Multi-Task | 3.657 | 0.806 |
| | Intraop | Individual | 0.967 | 0.829 |
| | | Multi-Task | 1.177 | 0.833 |
| | Postop | Individual | 2.209 | 0.852 |
| | | Multi-Task | 2.802 | 0.853 |
| Sepsis | Preop | Individual | 1.751 | 0.864 |
| | | Multi-Task | 3.922 | 0.868 |
| | Intraop | Individual | 1.533 | 0.780 |
| | | Multi-Task | 1.643 | 0.796 |
| | Postop | Individual | 1.885 | 0.875 |
| | | Multi-Task | 4.030 | 0.876 |
| AKI | Preop | Individual | 2.337 | 0.816 |
| | | Multi-Task | 4.541 | 0.819 |
| | Intraop | Individual | 0.955 | 0.730 |
| | | Multi-Task | 1.200 | 0.737 |
| | Postop | Individual | 3.146 | 0.826 |
| | | Multi-Task | 3.827 | 0.824 |
| Venous Thromboembolism | Preop | Individual | 1.781 | 0.819 |
| | | Multi-Task | 4.857 | 0.821 |
| | Intraop | Individual | 0.677 | 0.709 |
| | | Multi-Task | 1.173 | 0.735 |
| | Postop | Individual | 2.688 | 0.821 |
| | | Multi-Task | 5.000 | 0.827 |
| In-Hospital Mortality | Preop | Individual | 1.853 | 0.895 |
| | | Multi-Task | 3.675 | 0.893 |
| | Intraop | Individual | 0.727 | 0.858 |
| | | Multi-Task | 1.520 | 0.876 |
| | Postop | Individual | 1.820 | 0.903 |
| | | Multi-Task | 3.452 | 0.918 |

**Table 3. The 10 most influential features using integrated gradients aggregated over validation cohort.**

| Prolonged ICU Stay | Prolonged Mechanical Ventilation | Wound Complications | Neurological Complications | Cardiovascular Complications | Sepsis | Acute Kidney Injury | Venous Thromboembolism | In-Hospital Mortality |
|---|---|---|---|---|---|---|---|---|
| Peak inspiratory pressure (0.068) | Fraction of inspired oxygen (0.067) | Primary procedure (0.066) | Primary procedure (0.037) | Systolic blood pressure (0.068) | Heart rate (0.045) | Creatinine, serum (0.034) | Primary procedure (0.044) | Primary procedure (0.039) |
| Heart rate (0.063) | Peak inspiratory pressure (0.053) | Surgeon specialty (0.043) | Surgery type (0.035) | Peak inspiratory pressure (0.064) | Primary procedure (0.039) | Primary procedure (0.032) | Heart rate (0.036) | Minimum alveolar concentration (0.035) |
| Blood oxygen saturation (0.062) | Heart rate (0.053) | Attending surgeon (0.037) | Blood oxygen saturation (0.034) | Blood oxygen saturation (0.061) | Surgeon specialty (0.028) | Surgeon specialty (0.032) | Prothrombin time, serum (0.033) | Blood oxygen saturation (0.032) |
| Systolic blood pressure (0.048) | Blood oxygen saturation (0.044) | Surgery type (0.033) | Peak inspiratory pressure (0.033) | Heart rate (0.056) | Scheduled surgery room (0.025) | Attending surgeon (0.030) | Peak inspiratory pressure (0.026) | Peak inspiratory pressure (0.028) |
| Diastolic blood pressure (0.040) | Primary procedure (0.029) | Scheduled surgery room (0.026) | Erythrocytes, urine (0.032) | Diastolic blood pressure (0.045) | Blood oxygen saturation (0.024) | Peak inspiratory pressure (0.030) | Surgeon specialty (0.025) | Scheduled surgery room (0.026) |
| Primary procedure (0.031) | Respiratory rate (0.029) | ZIP code (0.025) | Minimum alveolar concentration (0.030) | Minimum alveolar concentration (0.033) | ZIP code (0.024) | Surgery type (0.028) | Blood oxygen saturation (0.025) | Erythrocytes, urine (0.026) |
| Fraction of inspired oxygen (0.030) | Scheduled surgery room (0.028) | Heart rate (0.025) | Scheduled surgery room (0.026) | Core temperature (0.030) | Surgery type (0.023) | Blood oxygen saturation (0.026) | Fraction of inspired oxygen (0.025) | Diastolic blood pressure (0.026) |
| Surgery duration (0.027) | Tidal volume (0.027) | Surgery duration (0.024) | Diastolic blood pressure (0.026) | Respiratory rate (0.021) | Erythrocytes, urine (0.023) | Heart rate (0.022) | Erythrocyte distribution width (0.024) | Fraction of inspired oxygen (0.022) |
| Mean arterial pressure (0.025) | Systolic blood pressure (0.023) | Albumin, serum (0.022) | Heart rate (0.025) | Primary procedure (0.021) | Peak inspiratory pressure (0.020) | Surgery duration (0.021) | Attending surgeon (0.023) | ZIP code (0.022) |
| Surgery type (0.023) | ZIP code (0.022) | Platelet mean volume (0.021) | Systolic blood pressure (0.024) | Surgery duration (0.020) | Attending surgeon (0.019) | Urea nitrogen, serum (0.020) | ZIP code (0.023) | Surgery type (0.021) |

# Dynamic Predictions of Postoperative Complications from Explainable, Uncertainty-Aware, and Multi-Task Deep Neural Networks

Benjamin Shickel, PhD [1,6], Tyler J. Loftus, MD [2,6], Matthew Ruppert, BS [1,3,6], Gilbert R. Upchurch Jr., MD [2], Tezcan Ozrazgat-Baslanti, PhD [1,3,6], Parisa Rashidi, PhD [1,4,5,6], Azra Bihorac, MD, MS [1,3,6,*]

[1] Department of Medicine, University of Florida, Gainesville, FL, 32611, USA
[2] Department of Surgery, University of Florida, Gainesville, FL, 32611, USA
[3] Precision and Intelligent Systems in Medicine (PRISMAp), University of Florida, Gainesville, FL, 32611, USA
[4] Department of Biomedical Engineering, University of Florida, Gainesville, FL, 32611, USA
[5] Intelligent Health Lab (i-Heal), University of Florida, Gainesville, FL, 32611, USA
[6] Intelligent Critical Care Center (IC$^3$), University of Florida, Gainesville, FL, 32611, USA

## SUPPLEMENTARY TABLES

**Supplementary Table S1. Summary of admission variables and prevalence of postoperative complications.**

| | **Development Cohort** (6/1/2014 - 11/26/2018) | **Validation Cohort** (11/27/2018 - 9/20/2020) |
|---|---|---|
| **Patients, n** | 38621 | 17621 |
| **Hospital encounters, n** | 47188 | 20293 |
| **Length of stay, days, median (25th, 75th)** | 4.1 (2.2, 7.9) | 4.3 (2.2, 8.4) |
| **Length of surgery, hours, median (25th, 75th)** | 3.1 (2.2, 4.6) | 3.2 (2.3, 4.7) |
| **Surgery lead time, hours, median (25th, 75th)** | 2.7 (2.0, 10.1) | 2.9 (2.1, 11.9) |
| **Admission Source, n (%)** | | |
| Non-Transfer | 39986 (84.7%) | 16976 (83.7%) |
| Transfer | 7202 (15.3%) | 3317 (16.3%) |
| **Emergent Admission, n (%)** | | |
| Non-Emergent | 30482 (64.6%) | 12802 (63.1%) |
| Emergent | 16706 (35.4%) | 7491 (36.9%) |
| **Admission Day, n (%)** | | |
| Weekday | 36811 (78.0%) | 15690 (77.3%) |
| Weekend | 10377 (22.0%) | 4603 (22.7%) |
| **Admission Time, n (%)** | | |
| Daytime | 25100 (53.2%) | 10587 (52.2%) |
| Nighttime | 22088 (46.8%) | 9706 (47.8%) |
| **Anesthesia Type, n (%)** | | |
| General | 42963 (91.0%) | 18741 (92.4%) |
| Local | 4225 (9.0%) | 1552 (7.6%) |
| **Admission Type, n (%)** | | |
| Medicine | 20893 (44.3%) | 8277 (40.8%) |
| Surgery | 17899 (37.9%) | 7806 (38.5%) |
| Other | 8396 (17.8%) | 4210 (20.7%) |
| **Surgery Type, n (%)** | | |
| Orthopaedic | 11831 (25.1%) | 5092 (25.1%) |
| Neurosurgery | 6757 (14.3%) | 3312 (16.3%) |
| Urologic | 4272 (9.1%) | 1459 (7.2%) |
| Vascular | 3731 (7.9%) | 1809 (8.9%) |
| Otolaryngology | 3261 (6.9%) | 1178 (5.8%) |
| Cardiothoracic | 3363 (7.1%) | 1038 (5.1%) |
| Gastrointestinal | 2935 (6.2%) | 1199 (5.9%) |
| Gynecologic | 2051 (4.3%) | 631 (3.1%) |
| Oncology | 1827 (3.9%) | 496 (2.4%) |
| Plastic | 1118 (2.4%) | 420 (2.1%) |
| Burn | 939 (2.0%) | 354 (1.7%) |

| | **Development Cohort** (6/1/2014 - 11/26/2018) | **Validation Cohort** (11/27/2018 - 9/20/2020) |
|---|---|---|
| Transplantation | 657 (1.4%) | 225 (1.1%) |
| Other | 4446 (9.4%) | 3080 (15.2%) |
| **Scheduled Postoperative Location, n (%)** | | |
| Non-ICU | 37143 (78.7%) | 18020 (88.8%) |
| ICU | 10045 (21.3%) | 2273 (11.2%) |
| **Postoperative Complications, n (%)** | | |
| Prolonged ICU Stay (> 2 Days) | 12980 (27.5%) | 6765 (33.3%) |
| Prolonged Mechanical Ventilation (> 2 Days) | 3512 (7.4%) | 1574 (7.8%) |
| Wound Complications | 6782 (14.4%) | 4347 (21.4%) |
| Neurological Complications | 7273 (15.4%) | 4107 (20.2%) |
| Cardiovascular Complications | 5655 (12.0%) | 3301 (16.3%) |
| Sepsis | 3445 (7.3%) | 1775 (8.7%) |
| Acute Kidney Injury | 6894 (14.6%) | 3438 (16.9%) |
| Venous Thromboembolism | 2008 (4.3%) | 1101 (5.4%) |
| In-Hospital Mortality | 788 (1.7%) | 321 (1.6%) |

## Supplementary Table S2. Summary of sociodemographic variables.

| | Development Cohort (6/1/2014 - 11/26/2018) | Validation Cohort (11/27/2018 - 9/20/2020) |
|---|---|---|
| **Patients, n** | 38621 | 17621 |
| **Hospital encounters, n** | 47188 | 20293 |
| **Age, years, median (25th, 75th)** | 59.0 (45.0, 69.0) | 61.0 (47.0, 71.0) |
| **Body mass index, median (25th, 75th)** | 28.1 (24.0, 33.3) | 28.1 (24.0, 33.3) |
| **Sex, n (%)** | | |
| Female | 23716 (50.3%) | 10005 (49.3%) |
| Male | 23472 (49.7%) | 10288 (50.7%) |
| **Ethnicity, n (%)** | | |
| Non-Hispanic | 44255 (93.8%) | 18817 (92.7%) |
| Hispanic | 2142 (4.5%) | 987 (4.9%) |
| Unknown | 791 (1.7%) | 489 (2.4%) |
| **Race, n (%)** | | |
| White | 37047 (78.5%) | 15942 (78.6%) |
| African American | 6562 (13.9%) | 2759 (13.6%) |
| Other/Unknown | 3579 (7.6%) | 1592 (7.8%) |
| **Language, n (%)** | | |
| English | 46342 (98.2%) | 19926 (98.2%) |
| Other | 846 (1.8%) | 367 (1.8%) |
| **Marital Status, n (%)** | | |
| Married | 22574 (47.8%) | 9586 (47.2%) |
| Single | 17040 (36.1%) | 7665 (37.8%) |
| Divorced | 7218 (15.3%) | 2902 (14.3%) |
| Unknown | 356 (0.8%) | 140 (0.7%) |
| **Smoking Status, n (%)** | | |
| Never | 20244 (42.9%) | 8986 (44.3%) |
| Former | 15948 (33.8%) | 6923 (34.1%) |
| Current | 8711 (18.5%) | 3485 (17.2%) |
| Unknown | 2285 (4.8%) | 899 (4.4%) |
| **Insurance, n (%)** | | |
| Medicare | 20859 (44.2%) | 9701 (47.8%) |
| Private | 14677 (31.1%) | 5740 (28.3%) |
| Medicaid | 7987 (16.9%) | 3023 (14.9%) |
| Uninsured | 3665 (7.8%) | 1829 (9.0%) |
| **Admission comorbidities, n (%)** | | |
| Myocardial infarction | 3253 (6.9%) | 1525 (7.5%) |
| Congestive heart failure | 6923 (14.7%) | 3505 (17.3%) |
| Peripheral vascular disease | 9569 (20.3%) | 4803 (23.7%) |
| Cerebrovascular disease | 7691 (16.3%) | 3537 (17.4%) |
| Chronic pulmonary disease | 14579 (30.9%) | 6715 (33.1%) |
| Metastatic carcinoma | 4474 (9.5%) | 1961 (9.7%) |
| Cancer | 13330 (28.2%) | 5366 (26.4%) |
| Liver disease | 6670 (14.1%) | 3132 (15.4%) |
| Diabetes | 10803 (22.9%) | 4806 (23.7%) |
| Hypertension | 29015 (61.5%) | 13346 (65.8%) |
| Hypothyroidism | 8196 (17.4%) | 3714 (18.3%) |
| Valvular disease | 5987 (12.7%) | 3316 (16.3%) |
| Coagulopathy | 6117 (13.0%) | 2709 (13.3%) |
| Obesity | 14855 (31.5%) | 8696 (42.9%) |
| Weight loss | 6445 (13.7%) | 3154 (15.5%) |
| Fluid/electrolyte disorders | 13266 (28.1%) | 7605 (37.5%) |
| Chronic anemia | 8762 (18.6%) | 5086 (25.1%) |
| Alcohol or drug abuse | 7102 (15.1%) | 3189 (15.7%) |
| Depression | 13001 (27.6%) | 6021 (29.7%) |
| **Unique diagnosis codes, n, median (25th, 75th)** | 40.0 (20.0, 88.0) | 47.0 (24.0, 108.0) |
| **Charlson comorbidity index, median (25th, 75th)** | 4.0 (2.0, 6.0) | 4.0 (2.0, 6.0) |
| **Neighborhood Characteristics, median (25th, 75th)** | | |
| Total population, n | 17583.0 (10722.5, 27063.0) | 17583.0 (10725.0, 27063.0) |

| | Development Cohort (6/1/2014 - 11/26/2018) | Validation Cohort (11/27/2018 - 9/20/2020) |
|---|---|---|
| Distance to hospital, km | 42.9 (22.1, 80.7) | 43.6 (22.3, 78.5) |
| Median income, dollars | 40528.0 (35244.0, 48493.0) | 40532.0 (35714.5, 48457.8) |
| Poverty rate | 19.1 (13.5, 25.0) | 18.8 (13.4, 24.9) |
| African American population proportion | 0.1 (0.0, 0.2) | 0.1 (0.0, 0.2) |
| Hispanic population proportion | 0.1 (0.0, 0.1) | 0.1 (0.0, 0.1) |
| **Rural/Urban, n (%)** | | |
| Urban | 30724 (65.1%) | 13119 (64.6%) |
| Rural | 16392 (34.7%) | 7132 (35.1%) |
| Unknown | 72 (0.2%) | 42 (0.2%) |

**Supplementary Table S3. Summary of medication and laboratory history variables.**

| | Development Cohort (6/1/2014 - 11/26/2018) | Validation Cohort (11/27/2018 - 9/20/2020) |
|---|---|---|
| **Patients, n** | 38621 | 17621 |
| **Hospital encounters, n** | 47188 | 20293 |
| **Received medications in past year, n (%)** | | |
| ACE Inhibitors | 4253 (9.0%) | 1677 (8.3%) |
| Aminoglycosides | 1631 (3.5%) | 801 (3.9%) |
| Antiemetics | 13282 (28.1%) | 5848 (28.8%) |
| Aspirin | 5902 (12.5%) | 2576 (12.7%) |
| Beta Blockers | 7164 (15.2%) | 3272 (16.1%) |
| Bicarbonates | 4672 (9.9%) | 2513 (12.4%) |
| Corticosteroids | 6692 (14.2%) | 3676 (18.1%) |
| Diuretics | 4731 (10.0%) | 1931 (9.5%) |
| NSAIDS | 6531 (13.8%) | 3239 (16.0%) |
| Vasopressors/Inotropes | 9187 (19.5%) | 4267 (21.0%) |
| Statins | 3803 (8.1%) | 1847 (9.1%) |
| Vancomycin | 4803 (10.2%) | 2445 (12.0%) |
| Nephrotoxic | 5338 (11.3%) | 2184 (10.8%) |
| Total medications, n, median (25th, 75th) | 0.0 (0.0, 3.0) | 0.0 (0.0, 3.0) |
| **Urea nitrogen/creatinine, ratio** | | |
| Minimum (0-7 days prior), median (25th, 75th) | 15.6 (11.9, 20.0) | 16.4 (12.5, 21.0) |
| Maximum (0-7 days prior), median (25th, 75th) | 17.6 (13.8, 22.8) | 18.6 (14.4, 23.8) |
| Average (0-7 days prior), median (25th, 75th) | 16.7 (13.0, 21.3) | 17.5 (13.7, 22.2) |
| Variance (0-7 days prior), median (25th, 75th) | 0.0 (0.0, 2.5) | 0.0 (0.0, 2.6) |
| Measurements (0-7 days prior), n, median (25th, 75th) | 4.0 (0.0, 8.0) | 5.0 (0.0, 10.0) |
| Minimum (8-365 days prior), median (25th, 75th) | 13.7 (10.2, 17.9) | 14.3 (10.8, 18.6) |
| Maximum (8-365 days prior), median (25th, 75th) | 19.8 (15.5, 25.6) | 21.0 (16.3, 27.3) |
| Average (8-365 days prior), median (25th, 75th) | 16.8 (13.4, 20.9) | 17.7 (14.2, 22.2) |
| Variance (8-365 days prior), median (25th, 75th) | 2.9 (0.0, 13.2) | 3.5 (0.0, 14.7) |
| Measurements (8-365 days prior), n, median (25th, 75th) | 4.0 (0.0, 15.0) | 5.0 (0.0, 19.0) |
| **Hemoglobin, g/dL** | | |
| Minimum (0-7 days prior), median (25th, 75th) | 12.6 (10.8, 14.0) | 12.5 (10.8, 13.9) |
| Maximum (0-7 days prior), median (25th, 75th) | 13.2 (11.8, 14.5) | 13.1 (11.7, 14.4) |
| Average (0-7 days prior), median (25th, 75th) | 12.9 (11.3, 14.2) | 12.8 (11.2, 14.1) |
| Variance (0-7 days prior), median (25th, 75th) | 0.4 (0.2, 1.1) | 0.4 (0.1, 1.0) |
| Measurements (0-7 days prior), n, median (25th, 75th) | 1.0 (0.0, 2.0) | 1.0 (0.0, 2.0) |
| Minimum (8-365 days prior), median (25th, 75th) | 12.1 (9.9, 13.7) | 12.1 (9.8, 13.6) |
| Maximum (8-365 days prior), median (25th, 75th) | 13.7 (12.6, 14.8) | 13.7 (12.5, 14.7) |
| Average (8-365 days prior), median (25th, 75th) | 12.7 (11.1, 14.0) | 12.7 (11.1, 13.9) |
| Variance (8-365 days prior), median (25th, 75th) | 0.9 (0.3, 1.9) | 0.8 (0.3, 1.7) |
| Measurements (8-365 days prior), n, median (25th, 75th) | 1.0 (0.0, 4.0) | 1.0 (0.0, 4.0) |
| **Leukocytes, thou/uL** | | |
| Minimum (0-7 days prior), median (25th, 75th) | 7.6 (6.0, 10.0) | 7.5 (5.8, 9.9) |
| Maximum (0-7 days prior), median (25th, 75th) | 8.7 (6.7, 12.1) | 8.6 (6.5, 11.9) |
| Average (0-7 days prior), median (25th, 75th) | 8.2 (6.4, 10.9) | 8.1 (6.3, 10.8) |
| Variance (0-7 days prior), median (25th, 75th) | 1.8 (0.5, 6.2) | 1.6 (0.4, 5.1) |
| Measurements (0-7 days prior), n, median (25th, 75th) | 1.0 (0.0, 2.0) | 1.0 (0.0, 2.0) |
| Minimum (8-365 days prior), median (25th, 75th) | 6.4 (5.0, 8.0) | 6.1 (4.8, 7.7) |
| Maximum (8-365 days prior), median (25th, 75th) | 9.0 (6.9, 12.6) | 8.9 (6.7, 12.7) |
| Average (8-365 days prior), median (25th, 75th) | 7.7 (6.2, 9.7) | 7.5 (6.0, 9.5) |
| Variance (8-365 days prior), median (25th, 75th) | 2.8 (0.8, 7.7) | 2.8 (0.8, 7.6) |
| Measurements (8-365 days prior), n, median (25th, 75th) | 1.0 (0.0, 3.0) | 1.0 (0.0, 3.0) |
| **Erythrocytes, million/uL** | | |
| Minimum (0-7 days prior), median (25th, 75th) | 4.2 (3.7, 4.7) | 4.2 (3.7, 4.6) |
| Maximum (0-7 days prior), median (25th, 75th) | 4.4 (4.0, 4.8) | 4.3 (3.9, 4.8) |
| Average (0-7 days prior), median (25th, 75th) | 4.3 (3.8, 4.7) | 4.3 (3.8, 4.7) |
| Variance (0-7 days prior), median (25th, 75th) | 0.0 (0.0, 0.1) | 0.0 (0.0, 0.1) |

| | Development Cohort (6/1/2014 - 11/26/2018) | Validation Cohort (11/27/2018 - 9/20/2020) |
|---|---|---|
| Measurements (0-7 days prior), n, median (25th, 75th) | 1.0 (0.0, 2.0) | 1.0 (0.0, 2.0) |
| Minimum (8-365 days prior), median (25th, 75th) | 4.1 (3.4, 4.6) | 4.1 (3.4, 4.5) |
| Maximum (8-365 days prior), median (25th, 75th) | 4.6 (4.2, 4.9) | 4.5 (4.2, 4.9) |
| Average (8-365 days prior), median (25th, 75th) | 4.3 (3.8, 4.7) | 4.3 (3.8, 4.7) |
| Variance (8-365 days prior), median (25th, 75th) | 0.1 (0.0, 0.2) | 0.1 (0.0, 0.2) |
| Measurements (8-365 days prior), n, median (25th, 75th) | 1.0 (0.0, 3.0) | 1.0 (0.0, 3.0) |
| **Hematocrit, %** | | |
| Minimum (0-7 days prior), median (25th, 75th) | 38.1 (33.1, 41.8) | 37.4 (32.5, 41.2) |
| Maximum (0-7 days prior), median (25th, 75th) | 39.8 (35.7, 43.1) | 39.2 (35.0, 42.7) |
| Average (0-7 days prior), median (25th, 75th) | 38.9 (34.5, 42.3) | 38.3 (33.8, 41.8) |
| Variance (0-7 days prior), median (25th, 75th) | 0.0 (0.0, 2.5) | 0.0 (0.0, 2.5) |
| Measurements (0-7 days prior), n, median (25th, 75th) | 2.0 (0.0, 4.0) | 2.0 (0.0, 4.0) |
| Minimum (8-365 days prior), median (25th, 75th) | 36.7 (30.3, 41.1) | 36.1 (29.6, 40.4) |
| Maximum (8-365 days prior), median (25th, 75th) | 41.3 (38.2, 44.4) | 40.7 (37.6, 43.6) |
| Average (8-365 days prior), median (25th, 75th) | 38.5 (34.2, 42.0) | 38.0 (33.4, 41.3) |
| Variance (8-365 days prior), median (25th, 75th) | 2.8 (0.0, 10.9) | 2.6 (0.0, 10.3) |
| Measurements (8-365 days prior), n, median (25th, 75th) | 2.0 (0.0, 6.0) | 2.0 (0.0, 6.0) |
| **Erythrocyte mean corpuscular volume, fL** | | |
| Minimum (0-7 days prior), median (25th, 75th) | 89.8 (85.9, 93.6) | 89.1 (85.1, 92.9) |
| Maximum (0-7 days prior), median (25th, 75th) | 90.6 (86.6, 94.4) | 89.8 (85.8, 93.6) |
| Average (0-7 days prior), median (25th, 75th) | 90.2 (86.3, 94.0) | 89.4 (85.4, 93.2) |
| Variance (0-7 days prior), median (25th, 75th) | 0.6 (0.2, 1.6) | 0.4 (0.1, 1.0) |
| Measurements (0-7 days prior), n, median (25th, 75th) | 1.0 (0.0, 2.0) | 1.0 (0.0, 2.0) |
| Minimum (8-365 days prior), median (25th, 75th) | 88.8 (84.7, 92.7) | 88.1 (83.7, 92.0) |
| Maximum (8-365 days prior), median (25th, 75th) | 91.6 (87.7, 95.7) | 90.7 (86.9, 94.7) |
| Average (8-365 days prior), median (25th, 75th) | 90.2 (86.4, 94.0) | 89.4 (85.4, 93.1) |
| Variance (8-365 days prior), median (25th, 75th) | 2.1 (0.7, 5.1) | 1.7 (0.6, 4.6) |
| Measurements (8-365 days prior), n, median (25th, 75th) | 1.0 (0.0, 3.0) | 1.0 (0.0, 3.0) |
| **Erythrocyte mean corpuscular hemoglobin concentration, g/dL** | | |
| Minimum (0-7 days prior), median (25th, 75th) | 32.9 (31.9, 33.8) | 33.3 (32.7, 34.0) |
| Maximum (0-7 days prior), median (25th, 75th) | 33.6 (32.8, 34.4) | 33.7 (33.0, 34.3) |
| Average (0-7 days prior), median (25th, 75th) | 33.3 (32.4, 34.0) | 33.5 (32.9, 34.1) |
| Variance (0-7 days prior), median (25th, 75th) | 0.2 (0.1, 0.4) | 0.2 (0.1, 0.3) |
| Measurements (0-7 days prior), n, median (25th, 75th) | 2.0 (0.0, 2.0) | 1.0 (0.0, 2.0) |
| Minimum (8-365 days prior), median (25th, 75th) | 32.4 (31.3, 33.4) | 33.0 (32.2, 33.7) |
| Maximum (8-365 days prior), median (25th, 75th) | 33.8 (33.0, 34.5) | 34.1 (33.3, 34.8) |
| Average (8-365 days prior), median (25th, 75th) | 33.1 (32.2, 33.8) | 33.5 (32.9, 34.1) |
| Variance (8-365 days prior), median (25th, 75th) | 0.4 (0.1, 0.7) | 0.3 (0.2, 0.6) |
| Measurements (8-365 days prior), n, median (25th, 75th) | 1.0 (0.0, 5.0) | 1.0 (0.0, 3.0) |
| **Erythrocyte mean corpuscular hemoglobin, pg** | | |
| Minimum (0-7 days prior), median (25th, 75th) | 29.9 (28.3, 31.3) | 30.0 (28.3, 31.4) |
| Maximum (0-7 days prior), median (25th, 75th) | 30.2 (28.6, 31.6) | 30.3 (28.6, 31.7) |
| Average (0-7 days prior), median (25th, 75th) | 30.0 (28.5, 31.4) | 30.1 (28.5, 31.5) |
| Variance (0-7 days prior), median (25th, 75th) | 0.1 (0.0, 0.3) | 0.1 (0.0, 0.2) |
| Measurements (0-7 days prior), n, median (25th, 75th) | 1.0 (0.0, 2.0) | 1.0 (0.0, 2.0) |
| Minimum (8-365 days prior), median (25th, 75th) | 29.4 (27.6, 30.9) | 29.5 (27.7, 31.1) |
| Maximum (8-365 days prior), median (25th, 75th) | 30.5 (28.9, 31.9) | 30.6 (29.0, 32.1) |
| Average (8-365 days prior), median (25th, 75th) | 29.9 (28.3, 31.3) | 30.1 (28.4, 31.5) |
| Variance (8-365 days prior), median (25th, 75th) | 0.3 (0.1, 0.7) | 0.3 (0.1, 0.7) |
| Measurements (8-365 days prior), n, median (25th, 75th) | 1.0 (0.0, 3.0) | 1.0 (0.0, 3.0) |
| **Erythrocyte distribution width, %** | | |
| Minimum (0-7 days prior), median (25th, 75th) | 14.1 (13.3, 15.2) | 14.1 (13.3, 15.4) |
| Maximum (0-7 days prior), median (25th, 75th) | 14.4 (13.5, 15.5) | 14.3 (13.5, 15.6) |
| Average (0-7 days prior), median (25th, 75th) | 14.2 (13.4, 15.3) | 14.2 (13.4, 15.5) |
| Variance (0-7 days prior), median (25th, 75th) | 0.1 (0.0, 0.3) | 0.0 (0.0, 0.1) |
| Measurements (0-7 days prior), n, median (25th, 75th) | 1.0 (0.0, 2.0) | 1.0 (0.0, 2.0) |
| Minimum (8-365 days prior), median (25th, 75th) | 13.8 (13.1, 14.7) | 13.8 (13.1, 14.7) |

| | Development Cohort (6/1/2014 - 11/26/2018) | Validation Cohort (11/27/2018 - 9/20/2020) |
|---|---|---|
| Maximum (8-365 days prior), median (25th, 75th) | 14.8 (13.8, 16.3) | 14.6 (13.7, 16.6) |
| Average (8-365 days prior), median (25th, 75th) | 14.3 (13.5, 15.4) | 14.3 (13.5, 15.6) |
| Variance (8-365 days prior), median (25th, 75th) | 0.4 (0.1, 0.9) | 0.2 (0.1, 0.9) |
| Measurements (8-365 days prior), n, median (25th, 75th) | 1.0 (0.0, 3.0) | 1.0 (0.0, 3.0) |
| **Platelets, thou/uL** | | |
| Minimum (0-7 days prior), median (25th, 75th) | 223.0 (176.0, 279.0) | 230.0 (181.0, 288.0) |
| Maximum (0-7 days prior), median (25th, 75th) | 241.0 (193.0, 301.0) | 248.0 (198.0, 310.0) |
| Average (0-7 days prior), median (25th, 75th) | 232.0 (185.3, 289.0) | 238.0 (190.0, 297.0) |
| Variance (0-7 days prior), median (25th, 75th) | 420.5 (126.2, 1195.0) | 387.9 (115.0, 1066.3) |
| Measurements (0-7 days prior), n, median (25th, 75th) | 1.0 (0.0, 2.0) | 1.0 (0.0, 2.0) |
| Minimum (8-365 days prior), median (25th, 75th) | 205.0 (157.0, 256.0) | 210.0 (158.0, 263.0) |
| Maximum (8-365 days prior), median (25th, 75th) | 264.0 (209.0, 341.0) | 273.0 (218.0, 351.0) |
| Average (8-365 days prior), median (25th, 75th) | 235.0 (190.0, 289.1) | 241.0 (194.0, 297.0) |
| Variance (8-365 days prior), median (25th, 75th) | 1006.4 (305.1, 3120.5) | 1018.2 (288.0, 3357.0) |
| Measurements (8-365 days prior), n, median (25th, 75th) | 1.0 (0.0, 3.0) | 1.0 (0.0, 4.0) |
| **Platelet mean volume, fL** | | |
| Minimum (0-7 days prior), median (25th, 75th) | 8.0 (7.3, 8.7) | 8.2 (7.6, 9.0) |
| Maximum (0-7 days prior), median (25th, 75th) | 8.4 (7.7, 9.2) | 8.5 (7.8, 9.3) |
| Average (0-7 days prior), median (25th, 75th) | 8.2 (7.5, 8.9) | 8.4 (7.7, 9.1) |
| Variance (0-7 days prior), median (25th, 75th) | 0.1 (0.0, 0.4) | 0.0 (0.0, 0.1) |
| Measurements (0-7 days prior), n, median (25th, 75th) | 1.0 (0.0, 2.0) | 1.0 (0.0, 2.0) |
| Minimum (8-365 days prior), median (25th, 75th) | 7.8 (7.1, 8.6) | 8.1 (7.4, 8.9) |
| Maximum (8-365 days prior), median (25th, 75th) | 8.8 (8.1, 9.8) | 9.1 (8.3, 10.1) |
| Average (8-365 days prior), median (25th, 75th) | 8.3 (7.6, 9.0) | 8.5 (7.9, 9.3) |
| Variance (8-365 days prior), median (25th, 75th) | 0.3 (0.1, 0.7) | 0.2 (0.1, 0.6) |
| Measurements (8-365 days prior), n, median (25th, 75th) | 1.0 (0.0, 3.0) | 1.0 (0.0, 3.0) |
| **Neutrophils, thou/uL** | | |
| Minimum (0-7 days prior), median (25th, 75th) | 4.0 (2.0, 11.0) | 3.0 (1.0, 7.0) |
| Maximum (0-7 days prior), median (25th, 75th) | 6.0 (2.0, 16.0) | 5.0 (2.0, 11.0) |
| Average (0-7 days prior), median (25th, 75th) | 5.6 (2.0, 13.0) | 4.0 (2.0, 9.0) |
| Variance (0-7 days prior), median (25th, 75th) | 12.5 (2.0, 59.8) | 6.6 (1.0, 36.3) |
| Measurements (0-7 days prior), n, median (25th, 75th) | 0.0 (0.0, 0.0) | 0.0 (0.0, 0.0) |
| Minimum (8-365 days prior), median (25th, 75th) | 2.0 (1.0, 6.0) | 1.0 (1.0, 3.0) |
| Maximum (8-365 days prior), median (25th, 75th) | 8.0 (3.0, 18.0) | 4.0 (2.0, 11.0) |
| Average (8-365 days prior), median (25th, 75th) | 5.0 (2.0, 11.0) | 3.0 (1.7, 6.0) |
| Variance (8-365 days prior), median (25th, 75th) | 21.1 (4.5, 65.7) | 6.2 (1.1, 23.0) |
| Measurements (8-365 days prior), n, median (25th, 75th) | 0.0 (0.0, 0.0) | 0.0 (0.0, 0.0) |
| **Glucose, serum, mg/dL** | | |
| Minimum (0-7 days prior), median (25th, 75th) | 100.0 (89.0, 119.0) | 101.0 (89.0, 120.0) |
| Maximum (0-7 days prior), median (25th, 75th) | 118.0 (98.0, 154.0) | 119.0 (100.0, 154.0) |
| Average (0-7 days prior), median (25th, 75th) | 110.0 (96.0, 134.7) | 110.0 (96.1, 135.0) |
| Variance (0-7 days prior), median (25th, 75th) | 8.3 (0.0, 266.7) | 16.0 (0.0, 280.3) |
| Measurements (0-7 days prior), n, median (25th, 75th) | 2.0 (0.0, 4.0) | 2.0 (1.0, 4.0) |
| Minimum (8-365 days prior), median (25th, 75th) | 90.0 (80.0, 102.0) | 91.0 (80.0, 103.0) |
| Maximum (8-365 days prior), median (25th, 75th) | 126.0 (100.0, 175.0) | 129.0 (102.0, 180.0) |
| Average (8-365 days prior), median (25th, 75th) | 107.3 (95.0, 127.0) | 108.1 (96.3, 128.0) |
| Variance (8-365 days prior), median (25th, 75th) | 128.3 (0.0, 571.9) | 144.8 (0.0, 630.4) |
| Measurements (8-365 days prior), n, median (25th, 75th) | 2.0 (0.0, 8.0) | 2.0 (0.0, 8.0) |
| **Urea nitrogen, serum, mg/dL** | | |
| Minimum (0-7 days prior), median (25th, 75th) | 14.0 (10.0, 19.0) | 15.0 (11.0, 20.0) |
| Maximum (0-7 days prior), median (25th, 75th) | 16.0 (12.0, 21.0) | 16.0 (12.0, 22.0) |
| Average (0-7 days prior), median (25th, 75th) | 14.7 (11.0, 19.8) | 15.6 (11.9, 21.0) |
| Variance (0-7 days prior), median (25th, 75th) | 0.0 (0.0, 2.7) | 0.0 (0.0, 2.9) |
| Measurements (0-7 days prior), n, median (25th, 75th) | 2.0 (0.0, 4.0) | 2.0 (0.0, 4.0) |
| Minimum (8-365 days prior), median (25th, 75th) | 12.0 (8.0, 16.0) | 13.0 (9.0, 17.0) |

| | Development Cohort (6/1/2014 - 11/26/2018) | Validation Cohort (11/27/2018 - 9/20/2020) |
|---|---|---|
| Maximum (8-365 days prior), median (25th, 75th) | 17.0 (13.0, 24.0) | 19.0 (14.0, 25.0) |
| Average (8-365 days prior), median (25th, 75th) | 14.8 (11.2, 19.0) | 15.5 (12.0, 20.0) |
| Variance (8-365 days prior), median (25th, 75th) | 3.0 (0.0, 12.6) | 3.3 (0.0, 14.4) |
| Measurements (8-365 days prior), n, median (25th, 75th) | 2.0 (0.0, 6.0) | 2.0 (0.0, 7.0) |
| **Creatinine, serum, mg/dL** | | |
| Minimum (0-7 days prior), median (25th, 75th) | 0.8 (0.7, 1.0) | 0.8 (0.7, 1.1) |
| Maximum (0-7 days prior), median (25th, 75th) | 0.9 (0.7, 1.1) | 0.9 (0.8, 1.1) |
| Average (0-7 days prior), median (25th, 75th) | 0.9 (0.7, 1.1) | 0.9 (0.7, 1.1) |
| Variance (0-7 days prior), median (25th, 75th) | 0.0 (0.0, 0.0) | 0.0 (0.0, 0.0) |
| Measurements (0-7 days prior), n, median (25th, 75th) | 2.0 (0.0, 4.0) | 2.0 (0.0, 4.0) |
| Minimum (8-365 days prior), median (25th, 75th) | 0.8 (0.6, 1.0) | 0.8 (0.6, 1.0) |
| Maximum (8-365 days prior), median (25th, 75th) | 0.9 (0.8, 1.2) | 1.0 (0.8, 1.2) |
| Average (8-365 days prior), median (25th, 75th) | 0.9 (0.7, 1.1) | 0.9 (0.7, 1.1) |
| Variance (8-365 days prior), median (25th, 75th) | 0.0 (0.0, 0.0) | 0.0 (0.0, 0.0) |
| Measurements (8-365 days prior), n, median (25th, 75th) | 2.0 (0.0, 6.0) | 2.0 (0.0, 8.0) |
| **Sodium, serum, mmol/L** | | |
| Minimum (0-7 days prior), median (25th, 75th) | 138.0 (136.0, 140.0) | 138.0 (135.0, 140.0) |
| Maximum (0-7 days prior), median (25th, 75th) | 140.0 (138.0, 141.0) | 139.0 (137.0, 141.0) |
| Average (0-7 days prior), median (25th, 75th) | 139.0 (137.0, 141.0) | 138.5 (136.3, 140.0) |
| Variance (0-7 days prior), median (25th, 75th) | 2.3 (1.0, 6.3) | 2.3 (0.8, 5.4) |
| Measurements (0-7 days prior), n, median (25th, 75th) | 1.0 (0.0, 2.0) | 1.0 (0.0, 2.0) |
| Minimum (8-365 days prior), median (25th, 75th) | 137.0 (134.0, 140.0) | 137.0 (134.0, 139.0) |
| Maximum (8-365 days prior), median (25th, 75th) | 141.0 (139.0, 143.0) | 141.0 (139.0, 142.0) |
| Average (8-365 days prior), median (25th, 75th) | 139.0 (137.3, 140.6) | 139.0 (137.0, 140.2) |
| Variance (8-365 days prior), median (25th, 75th) | 4.3 (2.0, 8.0) | 4.1 (2.0, 7.5) |
| Measurements (8-365 days prior), n, median (25th, 75th) | 1.0 (0.0, 3.0) | 1.0 (0.0, 4.0) |
| **Potassium, serum, mmol/L** | | |
| Minimum (0-7 days prior), median (25th, 75th) | 3.9 (3.6, 4.2) | 3.9 (3.6, 4.2) |
| Maximum (0-7 days prior), median (25th, 75th) | 4.2 (3.9, 4.5) | 4.1 (3.9, 4.5) |
| Average (0-7 days prior), median (25th, 75th) | 4.0 (3.8, 4.3) | 4.0 (3.8, 4.3) |
| Variance (0-7 days prior), median (25th, 75th) | 0.1 (0.0, 0.2) | 0.1 (0.0, 0.1) |
| Measurements (0-7 days prior), n, median (25th, 75th) | 1.0 (0.0, 2.0) | 1.0 (0.0, 2.0) |
| Minimum (8-365 days prior), median (25th, 75th) | 3.8 (3.5, 4.1) | 3.8 (3.4, 4.1) |
| Maximum (8-365 days prior), median (25th, 75th) | 4.4 (4.1, 4.8) | 4.4 (4.1, 4.8) |
| Average (8-365 days prior), median (25th, 75th) | 4.1 (3.9, 4.3) | 4.1 (3.9, 4.3) |
| Variance (8-365 days prior), median (25th, 75th) | 0.1 (0.0, 0.2) | 0.1 (0.0, 0.2) |
| Measurements (8-365 days prior), n, median (25th, 75th) | 1.0 (0.0, 3.0) | 1.0 (0.0, 4.0) |
| **Chloride, serum, mmol/L** | | |
| Minimum (0-7 days prior), median (25th, 75th) | 101.0 (98.0, 103.0) | 102.0 (100.0, 105.0) |
| Maximum (0-7 days prior), median (25th, 75th) | 103.0 (100.0, 105.0) | 104.0 (102.0, 106.0) |
| Average (0-7 days prior), median (25th, 75th) | 102.0 (99.0, 104.0) | 103.1 (101.0, 105.5) |
| Variance (0-7 days prior), median (25th, 75th) | 4.2 (1.4, 8.3) | 3.5 (1.2, 8.0) |
| Measurements (0-7 days prior), n, median (25th, 75th) | 1.0 (0.0, 2.0) | 1.0 (0.0, 2.0) |
| Minimum (8-365 days prior), median (25th, 75th) | 100.0 (97.0, 102.0) | 101.0 (98.0, 104.0) |
| Maximum (8-365 days prior), median (25th, 75th) | 104.0 (101.0, 107.0) | 105.0 (103.0, 108.0) |
| Average (8-365 days prior), median (25th, 75th) | 102.0 (100.0, 104.0) | 103.5 (101.5, 105.1) |
| Variance (8-365 days prior), median (25th, 75th) | 5.7 (2.3, 11.3) | 4.9 (2.0, 10.3) |
| Measurements (8-365 days prior), n, median (25th, 75th) | 1.0 (0.0, 3.0) | 1.0 (0.0, 4.0) |
| **Carbon dioxide, serum, mmol/L** | | |
| Minimum (0-7 days prior), median (25th, 75th) | 24.0 (22.0, 26.0) | 25.0 (22.0, 27.0) |
| Maximum (0-7 days prior), median (25th, 75th) | 26.0 (24.0, 28.0) | 26.0 (24.0, 28.0) |
| Average (0-7 days prior), median (25th, 75th) | 25.0 (23.0, 27.0) | 25.5 (23.5, 27.7) |
| Variance (0-7 days prior), median (25th, 75th) | 2.7 (1.0, 6.3) | 2.9 (1.0, 6.4) |
| Measurements (0-7 days prior), n, median (25th, 75th) | 1.0 (0.0, 2.0) | 1.0 (0.0, 2.0) |
| Minimum (8-365 days prior), median (25th, 75th) | 23.0 (21.0, 26.0) | 24.0 (21.0, 27.0) |
| Maximum (8-365 days prior), median (25th, 75th) | 27.0 (25.0, 29.0) | 28.0 (26.0, 30.0) |
| Average (8-365 days prior), median (25th, 75th) | 25.2 (23.6, 27.0) | 26.0 (24.1, 27.9) |

| | Development Cohort (6/1/2014 - 11/26/2018) | Validation Cohort (11/27/2018 - 9/20/2020) |
|---|---|---|
| Variance (8-365 days prior), median (25th, 75th) | 4.5 (2.0, 8.0) | 4.5 (2.2, 8.0) |
| Measurements (8-365 days prior), n, median (25th, 75th) | 1.0 (0.0, 3.0) | 1.0 (0.0, 4.0) |
| **Lactate, serum, mmol/L** | | |
| Minimum (0-7 days prior), median (25th, 75th) | 1.3 (0.9, 1.8) | 1.2 (0.9, 1.8) |
| Maximum (0-7 days prior), median (25th, 75th) | 1.6 (1.1, 2.6) | 1.7 (1.2, 2.6) |
| Average (0-7 days prior), median (25th, 75th) | 1.5 (1.1, 2.2) | 1.5 (1.1, 2.1) |
| Variance (0-7 days prior), median (25th, 75th) | 0.2 (0.0, 0.7) | 0.1 (0.0, 0.5) |
| Measurements (0-7 days prior), n, median (25th, 75th) | 0.0 (0.0, 0.0) | 0.0 (0.0, 0.0) |
| Minimum (8-365 days prior), median (25th, 75th) | 0.9 (0.7, 1.3) | 0.9 (0.7, 1.2) |
| Maximum (8-365 days prior), median (25th, 75th) | 1.7 (1.2, 2.7) | 1.8 (1.2, 2.9) |
| Average (8-365 days prior), median (25th, 75th) | 1.3 (1.0, 1.8) | 1.4 (1.0, 1.8) |
| Variance (8-365 days prior), median (25th, 75th) | 0.3 (0.1, 0.8) | 0.2 (0.1, 0.8) |
| Measurements (8-365 days prior), n, median (25th, 75th) | 0.0 (0.0, 0.0) | 0.0 (0.0, 0.0) |
| **Calcium, serum, mg/dL** | | |
| Minimum (0-7 days prior), median (25th, 75th) | 9.1 (8.6, 9.6) | 9.1 (8.6, 9.6) |
| Maximum (0-7 days prior), median (25th, 75th) | 9.4 (9.0, 9.7) | 9.4 (9.0, 9.7) |
| Average (0-7 days prior), median (25th, 75th) | 9.2 (8.8, 9.6) | 9.2 (8.8, 9.6) |
| Variance (0-7 days prior), median (25th, 75th) | 0.0 (0.0, 0.1) | 0.0 (0.0, 0.1) |
| Measurements (0-7 days prior), n, median (25th, 75th) | 2.0 (0.0, 4.0) | 2.0 (0.0, 4.0) |
| Minimum (8-365 days prior), median (25th, 75th) | 9.0 (8.3, 9.5) | 9.0 (8.2, 9.5) |
| Maximum (8-365 days prior), median (25th, 75th) | 9.6 (9.3, 9.9) | 9.6 (9.3, 9.9) |
| Average (8-365 days prior), median (25th, 75th) | 9.3 (8.9, 9.6) | 9.3 (8.9, 9.6) |
| Variance (8-365 days prior), median (25th, 75th) | 0.1 (0.0, 0.2) | 0.1 (0.0, 0.2) |
| Measurements (8-365 days prior), n, median (25th, 75th) | 2.0 (0.0, 6.0) | 2.0 (0.0, 6.0) |
| **Anion gap, serum, mmol/L** | | |
| Minimum (0-7 days prior), median (25th, 75th) | 11.0 (9.0, 14.0) | 9.0 (7.0, 10.0) |
| Maximum (0-7 days prior), median (25th, 75th) | 13.0 (11.0, 16.0) | 10.0 (9.0, 13.0) |
| Average (0-7 days prior), median (25th, 75th) | 12.0 (10.0, 14.7) | 9.7 (8.0, 11.0) |
| Variance (0-7 days prior), median (25th, 75th) | 3.7 (1.3, 8.3) | 3.0 (1.1, 8.0) |
| Measurements (0-7 days prior), n, median (25th, 75th) | 1.0 (0.0, 2.0) | 1.0 (0.0, 2.0) |
| Minimum (8-365 days prior), median (25th, 75th) | 10.0 (7.0, 12.0) | 8.0 (6.0, 9.0) |
| Maximum (8-365 days prior), median (25th, 75th) | 14.0 (11.0, 17.0) | 11.0 (9.0, 13.0) |
| Average (8-365 days prior), median (25th, 75th) | 12.0 (9.7, 14.0) | 9.0 (8.0, 10.2) |
| Variance (8-365 days prior), median (25th, 75th) | 4.8 (2.3, 9.3) | 3.5 (2.0, 6.3) |
| Measurements (8-365 days prior), n, median (25th, 75th) | 0.0 (0.0, 3.0) | 1.0 (0.0, 3.0) |
| **Alanine, serum, U/L** | | |
| Minimum (0-7 days prior), median (25th, 75th) | 18.0 (12.0, 28.0) | 16.0 (11.0, 26.0) |
| Maximum (0-7 days prior), median (25th, 75th) | 19.0 (12.0, 31.0) | 17.0 (11.0, 28.0) |
| Average (0-7 days prior), median (25th, 75th) | 18.0 (12.0, 30.0) | 16.0 (11.0, 27.0) |
| Variance (0-7 days prior), median (25th, 75th) | 8.0 (2.0, 95.5) | 8.0 (2.0, 60.5) |
| Measurements (0-7 days prior), n, median (25th, 75th) | 0.0 (0.0, 1.0) | 0.0 (0.0, 1.0) |
| Minimum (8-365 days prior), median (25th, 75th) | 15.0 (10.0, 22.0) | 13.0 (9.0, 20.0) |
| Maximum (8-365 days prior), median (25th, 75th) | 21.0 (14.0, 37.0) | 20.0 (13.0, 35.0) |
| Average (8-365 days prior), median (25th, 75th) | 18.5 (13.0, 28.9) | 17.0 (12.0, 26.1) |
| Variance (8-365 days prior), median (25th, 75th) | 26.1 (5.3, 162.0) | 22.7 (4.5, 135.9) |
| Measurements (8-365 days prior), n, median (25th, 75th) | 0.0 (0.0, 1.0) | 0.0 (0.0, 1.0) |
| **Albumin, serum, g/dL** | | |
| Minimum (0-7 days prior), median (25th, 75th) | 3.9 (3.3, 4.3) | 3.8 (3.2, 4.2) |
| Maximum (0-7 days prior), median (25th, 75th) | 4.0 (3.4, 4.3) | 3.9 (3.4, 4.3) |
| Average (0-7 days prior), median (25th, 75th) | 3.9 (3.3, 4.3) | 3.9 (3.3, 4.2) |
| Variance (0-7 days prior), median (25th, 75th) | 0.1 (0.0, 0.1) | 0.0 (0.0, 0.1) |
| Measurements (0-7 days prior), n, median (25th, 75th) | 0.0 (0.0, 1.0) | 0.0 (0.0, 1.0) |
| Minimum (8-365 days prior), median (25th, 75th) | 3.9 (3.2, 4.2) | 3.8 (3.2, 4.2) |
| Maximum (8-365 days prior), median (25th, 75th) | 4.2 (3.9, 4.5) | 4.2 (3.9, 4.4) |
| Average (8-365 days prior), median (25th, 75th) | 4.0 (3.6, 4.3) | 4.0 (3.6, 4.3) |
| Variance (8-365 days prior), median (25th, 75th) | 0.1 (0.0, 0.2) | 0.1 (0.0, 0.2) |
| Measurements (8-365 days prior), n, median (25th, 75th) | 0.0 (0.0, 1.0) | 0.0 (0.0, 1.0) |

| | Development Cohort (6/1/2014 - 11/26/2018) | Validation Cohort (11/27/2018 - 9/20/2020) |
|---|---|---|
| **Asparate, serum, U/L** | | |
| Minimum (0-7 days prior), median (25th, 75th) | 21.0 (16.0, 30.0) | 21.0 (16.0, 29.0) |
| Maximum (0-7 days prior), median (25th, 75th) | 22.0 (17.0, 34.0) | 22.0 (17.0, 33.0) |
| Average (0-7 days prior), median (25th, 75th) | 22.0 (16.0, 32.0) | 22.0 (17.0, 31.4) |
| Variance (0-7 days prior), median (25th, 75th) | 21.0 (4.5, 145.6) | 18.0 (2.7, 113.5) |
| Measurements (0-7 days prior), n, median (25th, 75th) | 0.0 (0.0, 1.0) | 0.0 (0.0, 1.0) |
| Minimum (8-365 days prior), median (25th, 75th) | 18.0 (14.0, 23.0) | 17.0 (14.0, 22.0) |
| Maximum (8-365 days prior), median (25th, 75th) | 24.0 (18.0, 37.0) | 23.0 (18.0, 37.0) |
| Average (8-365 days prior), median (25th, 75th) | 21.0 (16.1, 29.0) | 20.6 (16.5, 28.0) |
| Variance (8-365 days prior), median (25th, 75th) | 27.0 (6.7, 140.7) | 24.5 (5.0, 133.0) |
| Measurements (8-365 days prior), n, median (25th, 75th) | 0.0 (0.0, 1.0) | 0.0 (0.0, 1.0) |
| **Bilirubin, serum, mg/dL** | | |
| Minimum (0-7 days prior), median (25th, 75th) | 0.2 (0.1, 0.2) | 0.1 (0.1, 0.2) |
| Maximum (0-7 days prior), median (25th, 75th) | 0.2 (0.2, 0.2) | 0.1 (0.1, 0.2) |
| Average (0-7 days prior), median (25th, 75th) | 0.2 (0.2, 0.2) | 0.1 (0.1, 0.2) |
| Variance (0-7 days prior), median (25th, 75th) | 0.0 (0.0, 0.0) | 0.0 (0.0, 0.0) |
| Measurements (0-7 days prior), n, median (25th, 75th) | 0.0 (0.0, 0.0) | 0.0 (0.0, 0.0) |
| Minimum (8-365 days prior), median (25th, 75th) | 0.2 (0.1, 0.2) | 0.1 (0.1, 0.2) |
| Maximum (8-365 days prior), median (25th, 75th) | 0.2 (0.2, 0.2) | 0.2 (0.1, 0.3) |
| Average (8-365 days prior), median (25th, 75th) | 0.2 (0.2, 0.2) | 0.1 (0.1, 0.2) |
| Variance (8-365 days prior), median (25th, 75th) | 0.0 (0.0, 0.0) | 0.0 (0.0, 0.0) |
| Measurements (8-365 days prior), n, median (25th, 75th) | 0.0 (0.0, 0.0) | 0.0 (0.0, 0.0) |
| **C-reactive protein, serum, mg/L** | | |
| Minimum (0-7 days prior), median (25th, 75th) | 36.1 (6.6, 107.3) | 32.6 (7.0, 110.7) |
| Maximum (0-7 days prior), median (25th, 75th) | 39.9 (7.1, 113.9) | 35.8 (7.4, 117.6) |
| Average (0-7 days prior), median (25th, 75th) | 38.7 (7.0, 111.2) | 35.0 (7.2, 114.3) |
| Variance (0-7 days prior), median (25th, 75th) | 0.0 (0.0, 0.0) | 0.0 (0.0, 0.0) |
| Measurements (0-7 days prior), n, median (25th, 75th) | 0.0 (0.0, 0.0) | 0.0 (0.0, 0.0) |
| Minimum (8-365 days prior), median (25th, 75th) | 9.4 (2.9, 45.4) | 9.7 (2.8, 34.7) |
| Maximum (8-365 days prior), median (25th, 75th) | 33.2 (5.3, 113.3) | 29.4 (6.0, 118.2) |
| Average (8-365 days prior), median (25th, 75th) | 23.3 (4.8, 77.3) | 22.9 (5.3, 73.0) |
| Variance (8-365 days prior), median (25th, 75th) | 0.0 (0.0, 375.2) | 0.0 (0.0, 407.6) |
| Measurements (8-365 days prior), n, median (25th, 75th) | 0.0 (0.0, 0.0) | 0.0 (0.0, 0.0) |
| **Prothrombin time, serum, INR** | | |
| Minimum (0-7 days prior), median (25th, 75th) | 1.1 (1.0, 1.2) | 1.0 (1.0, 1.2) |
| Maximum (0-7 days prior), median (25th, 75th) | 1.1 (1.0, 1.2) | 1.1 (1.0, 1.2) |
| Average (0-7 days prior), median (25th, 75th) | 1.1 (1.0, 1.2) | 1.1 (1.0, 1.2) |
| Variance (0-7 days prior), median (25th, 75th) | 0.0 (0.0, 0.0) | 0.0 (0.0, 0.0) |
| Measurements (0-7 days prior), n, median (25th, 75th) | 0.0 (0.0, 1.0) | 0.0 (0.0, 1.0) |
| Minimum (8-365 days prior), median (25th, 75th) | 1.0 (1.0, 1.1) | 1.0 (1.0, 1.1) |
| Maximum (8-365 days prior), median (25th, 75th) | 1.1 (1.0, 1.2) | 1.1 (1.0, 1.3) |
| Average (8-365 days prior), median (25th, 75th) | 1.1 (1.0, 1.2) | 1.1 (1.0, 1.2) |
| Variance (8-365 days prior), median (25th, 75th) | 0.0 (0.0, 0.0) | 0.0 (0.0, 0.0) |
| Measurements (8-365 days prior), n, median (25th, 75th) | 0.0 (0.0, 1.0) | 0.0 (0.0, 1.0) |
| **Erythrocyte sedimentation rate, serum, mm/h** | | |
| Minimum (0-7 days prior), median (25th, 75th) | 44.0 (21.0, 80.0) | 11.0 (7.5, 39.0) |
| Maximum (0-7 days prior), median (25th, 75th) | 46.0 (22.0, 84.0) | 11.0 (7.5, 39.0) |
| Average (0-7 days prior), median (25th, 75th) | 45.0 (22.0, 82.0) | 11.0 (7.5, 39.0) |
| Variance (0-7 days prior), median (25th, 75th) | 72.0 (12.5, 378.2) | 0.0 (0.0, 0.0) |
| Measurements (0-7 days prior), n, median (25th, 75th) | 0.0 (0.0, 0.0) | 0.0 (0.0, 0.0) |
| Minimum (8-365 days prior), median (25th, 75th) | 27.0 (11.0, 54.0) | 11.0 (6.0, 28.0) |
| Maximum (8-365 days prior), median (25th, 75th) | 37.0 (17.0, 74.2) | 14.0 (6.0, 31.0) |
| Average (8-365 days prior), median (25th, 75th) | 33.0 (15.0, 64.0) | 14.0 (6.0, 29.0) |
| Variance (8-365 days prior), median (25th, 75th) | 220.5 (34.0, 722.0) | 50.0 (11.7, 331.6) |
| Measurements (8-365 days prior), n, median (25th, 75th) | 0.0 (0.0, 0.0) | 0.0 (0.0, 0.0) |
| **Troponin I, serum, ng/mL** | | |
| Minimum (0-7 days prior), median (25th, 75th) | 0.0 (0.0, 0.1) | 0.1 (0.1, 0.1) |

| | Development Cohort (6/1/2014 - 11/26/2018) | Validation Cohort (11/27/2018 - 9/20/2020) |
|---|---|---|
| Maximum (0-7 days prior), median (25th, 75th) | 0.0 (0.0, 0.1) | 0.1 (0.1, 0.1) |
| Average (0-7 days prior), median (25th, 75th) | 0.0 (0.0, 0.1) | 0.1 (0.1, 0.1) |
| Variance (0-7 days prior), median (25th, 75th) | 0.0 (0.0, 0.0) | 0.0 (0.0, 0.0) |
| Measurements (0-7 days prior), n, median (25th, 75th) | 0.0 (0.0, 0.0) | 0.0 (0.0, 0.0) |
| Minimum (8-365 days prior), median (25th, 75th) | 0.0 (0.0, 0.1) | 0.1 (0.1, 0.1) |
| Maximum (8-365 days prior), median (25th, 75th) | 0.0 (0.0, 0.1) | 0.1 (0.1, 0.1) |
| Average (8-365 days prior), median (25th, 75th) | 0.0 (0.0, 0.1) | 0.1 (0.1, 0.1) |
| Variance (8-365 days prior), median (25th, 75th) | 0.0 (0.0, 0.0) | 0.0 (0.0, 0.0) |
| Measurements (8-365 days prior), n, median (25th, 75th) | 0.0 (0.0, 0.0) | 0.0 (0.0, 0.0) |
| **Troponin T, serum, ng/mL** | | |
| Minimum (0-7 days prior), median (25th, 75th) | 0.0 (0.0, 0.0) | 0.0 (0.0, 0.0) |
| Maximum (0-7 days prior), median (25th, 75th) | 0.0 (0.0, 0.0) | 0.0 (0.0, 0.0) |
| Average (0-7 days prior), median (25th, 75th) | 0.0 (0.0, 0.0) | 0.0 (0.0, 0.0) |
| Variance (0-7 days prior), median (25th, 75th) | 0.0 (0.0, 0.0) | 0.0 (0.0, 0.0) |
| Measurements (0-7 days prior), n, median (25th, 75th) | 0.0 (0.0, 0.0) | 0.0 (0.0, 0.0) |
| Minimum (8-365 days prior), median (25th, 75th) | 0.0 (0.0, 0.0) | 0.0 (0.0, 0.0) |
| Maximum (8-365 days prior), median (25th, 75th) | 0.0 (0.0, 0.0) | 0.0 (0.0, 0.0) |
| Average (8-365 days prior), median (25th, 75th) | 0.0 (0.0, 0.0) | 0.0 (0.0, 0.0) |
| Variance (8-365 days prior), median (25th, 75th) | 0.0 (0.0, 0.0) | 0.0 (0.0, 0.0) |
| Measurements (8-365 days prior), n, median (25th, 75th) | 0.0 (0.0, 0.0) | 0.0 (0.0, 0.0) |
| **Erythrocytes, urine, /HPF** | | |
| Large (0-7 days prior) | 1109 (2.4%) | 435 (2.1%) |
| Moderate (0-7 days prior) | 442 (0.9%) | 256 (1.3%) |
| Small (0-7 days prior) | 1180 (2.5%) | 241 (1.2%) |
| Negative (0-7 days prior) | 6973 (14.8%) | 3325 (16.4%) |
| Missing (0-7 days prior) | 37484 (79.4%) | 16036 (79.0%) |
| Count (0-7 days prior), n, median (25th, 75th) | 0.0 (0.0, 0.0) | 0.0 (0.0, 0.0) |
| Large (8-365 days prior) | 970 (2.1%) | 347 (1.7%) |
| Moderate (8-365 days prior) | 259 (0.5%) | 172 (0.8%) |
| Small (8-365 days prior) | 812 (1.7%) | 129 (0.6%) |
| Negative (8-365 days prior) | 8740 (18.5%) | 3982 (19.6%) |
| Missing (8-365 days prior) | 36407 (77.2%) | 15663 (77.2%) |
| Count (8-365 days prior), n, median (25th, 75th) | 0.0 (0.0, 0.0) | 0.0 (0.0, 0.0) |
| Large (0-365 days prior) | 1674 (3.5%) | 598 (2.9%) |
| Moderate (0-365 days prior) | 527 (1.1%) | 312 (1.5%) |
| Small (0-365 days prior) | 1488 (3.2%) | 284 (1.4%) |
| Negative (0-365 days prior) | 14276 (30.3%) | 6626 (32.7%) |
| Missing (0-365 days prior) | 29223 (61.9%) | 12473 (61.5%) |
| Count (0-365 days prior), n, median (25th, 75th) | 0.0 (0.0, 1.0) | 0.0 (0.0, 2.0) |
| **Protein, urine, mg/dL** | | |
| Large (0-365 days prior) | 965 (2.0%) | 418 (2.1%) |
| Moderate (0-365 days prior) | 4937 (10.5%) | 2235 (11.0%) |
| Small (0-365 days prior) | 1281 (2.7%) | 181 (0.9%) |
| Negative (0-365 days prior) | 14757 (31.3%) | 5915 (29.1%) |
| Missing (0-365 days prior) | 25248 (53.5%) | 11544 (56.9%) |
| Count (0-7 days prior), n, median (25th, 75th) | 0.0 (0.0, 0.0) | 0.0 (0.0, 0.0) |
| Count (8-365 days prior), n, median (25th, 75th) | 0.0 (0.0, 1.0) | 0.0 (0.0, 1.0) |
| Count (0-365 days prior), n, median (25th, 75th) | 0.0 (0.0, 1.0) | 0.0 (0.0, 1.0) |
| **Glucose, urine, mg/dL** | | |
| Large (0-7 days prior) | 242 (0.5%) | 19 (0.1%) |
| Moderate (0-7 days prior) | 345 (0.7%) | 254 (1.3%) |
| Small (0-7 days prior) | 649 (1.4%) | 270 (1.3%) |
| Negative (0-7 days prior) | 9893 (21.0%) | 3887 (19.2%) |
| Missing (0-7 days prior) | 36059 (76.4%) | 15863 (78.2%) |
| Count (0-7 days prior), n, median (25th, 75th) | 0.0 (0.0, 0.0) | 0.0 (0.0, 0.0) |
| Large (8-365 days prior) | 325 (0.7%) | 58 (0.3%) |
| Moderate (8-365 days prior) | 289 (0.6%) | 226 (1.1%) |

| | **Development Cohort** (6/1/2014 - 11/26/2018) | **Validation Cohort** (11/27/2018 - 9/20/2020) |
|---|---|---|
| Small (8-365 days prior) | 367 (0.8%) | 146 (0.7%) |
| Negative (8-365 days prior) | 13289 (28.2%) | 5116 (25.2%) |
| Missing (8-365 days prior) | 32918 (69.8%) | 14747 (72.7%) |
| Count (8-365 days prior), n, median (25th, 75th) | 0.0 (0.0, 1.0) | 0.0 (0.0, 1.0) |
| **Hemoglobin, urine, /HPF** | | |
| Count (0-7 days prior), n, median (25th, 75th) | 0.0 (0.0, 0.0) | 0.0 (0.0, 0.0) |
| Count (8-365 days prior), n, median (25th, 75th) | 0.0 (0.0, 0.0) | 0.0 (0.0, 0.0) |

## Supplementary Table S4. Summary of intraoperative physiological variables.

| | Development Cohort (6/1/2014 - 11/26/2018) | Validation Cohort (11/27/2018 - 9/20/2020) |
|---|---|---|
| **Patients, n** | 38621 | 17621 |
| **Hospital encounters, n** | 47188 | 20293 |
| **Systolic blood pressure** | | |
| Measured value, mmHg, median (25th, 75th) | 113.0 (100.0, 128.0) | 114.0 (101.0, 130.0) |
| Total measurements, n, (25th, 75th) | 66.0 (34.0, 138.0) | 67.0 (29.0, 157.0) |
| Frequency, n/hour, (25th, 75th) | 20.1 (17.7, 46.3) | 20.0 (17.1, 48.9) |
| Encounters missing, n (%) | 36 (0.1%) | 14 (0.1%) |
| **Diastolic blood pressure** | | |
| Measured value, mmHg, median (25th, 75th) | 62.0 (53.2, 71.0) | 63.0 (54.0, 73.0) |
| Total measurements, n, (25th, 75th) | 66.0 (34.0, 138.0) | 67.0 (29.0, 157.0) |
| Frequency, n/hour, (25th, 75th) | 20.1 (17.7, 46.3) | 20.0 (17.1, 48.9) |
| Encounters missing, n (%) | 36 (0.1%) | 14 (0.1%) |
| **Mean arterial pressure** | | |
| Measured value, mmHg, median (25th, 75th) | 78.0 (69.0, 89.0) | 80.5 (71.0, 92.0) |
| Total measurements, n, (25th, 75th) | 66.0 (34.0, 139.0) | 67.0 (29.0, 158.0) |
| Frequency, n/hour, (25th, 75th) | 20.1 (17.7, 46.6) | 20.0 (17.1, 49.2) |
| Encounters missing, n (%) | 36 (0.1%) | 14 (0.1%) |
| **Heart rate** | | |
| Measured value, bpm, median (25th, 75th) | 75.5 (65.5, 87.0) | 75.5 (65.5, 87.0) |
| Total measurements, n, (25th, 75th) | 161.0 (88.0, 247.0) | 157.0 (75.0, 245.0) |
| Frequency, n/hour, (25th, 75th) | 54.9 (49.8, 56.9) | 54.1 (46.8, 56.4) |
| Encounters missing, n (%) | 38 (0.1%) | 10 (0.0%) |
| **Oxygen saturation (SpO2)** | | |
| Measured value, %, median (25th, 75th) | 99.0 (97.4, 100.0) | 99.0 (97.1, 100.0) |
| Total measurements, n, (25th, 75th) | 162.0 (86.0, 254.0) | 159.0 (73.0, 258.0) |
| Frequency, n/hour, (25th, 75th) | 54.5 (47.5, 56.9) | 53.7 (43.1, 56.6) |
| Encounters missing, n (%) | 27 (0.1%) | 8 (0.0%) |
| **Fraction of inspired oxygen (FiO2)** | | |
| Measured value, %, median (25th, 75th) | 40.0 (40.0, 40.0) | 40.0 (40.0, 40.0) |
| Total measurements, n, (25th, 75th) | 171.0 (95.0, 271.0) | 171.0 (83.0, 276.0) |
| Frequency, n/hour, (25th, 75th) | 57.1 (53.5, 58.6) | 56.5 (51.8, 58.5) |
| Encounters missing, n (%) | 27 (0.1%) | 8 (0.0%) |
| **End-tidal carbon dioxide (EtCO2)** | | |
| Measured value, mmHg, median (25th, 75th) | 34.0 (32.0, 37.0) | 35.0 (32.0, 38.0) |
| Total measurements, n, (25th, 75th) | 130.0 (2.0, 224.0) | 0.0 (0.0, 152.0) |
| Frequency, n/hour, (25th, 75th) | 49.1 (0.5, 53.5) | 0.0 (0.0, 47.9) |
| Encounters missing, n (%) | 27 (0.1%) | 8 (0.0%) |
| **Tidal volume** | | |
| Measured value, mL, median (25th, 75th) | 442.0 (361.0, 508.0) | 467.0 (388.0, 527.0) |
| Total measurements, n, (25th, 75th) | 154.0 (78.0, 248.0) | 136.0 (0.0, 238.0) |
| Frequency, n/hour, (25th, 75th) | 51.9 (43.5, 55.2) | 49.6 (0.0, 54.3) |
| Encounters missing, n (%) | 27 (0.1%) | 8 (0.0%) |
| **Respiration rate** | | |
| Measured value, breaths/minute, median (25th, 75th) | 10.0 (8.0, 12.0) | 12.0 (10.0, 14.0) |
| Total measurements, n, (25th, 75th) | 167.0 (91.0, 265.0) | 63.0 (1.0, 185.0) |
| Frequency, n/hour, (25th, 75th) | 55.9 (51.1, 58.0) | 43.4 (0.2, 54.6) |
| Encounters missing, n (%) | 27 (0.1%) | 8 (0.0%) |
| **Peak inspiratory pressure** | | |
| Measured value, mmHg, median (25th, 75th) | 18.0 (13.0, 23.0) | 18.0 (13.0, 22.0) |
| Total measurements, n, (25th, 75th) | 168.0 (91.0, 266.0) | 153.0 (43.0, 259.0) |
| Frequency, n/hour, (25th, 75th) | 56.1 (51.8, 58.0) | 54.8 (36.5, 57.6) |
| Encounters missing, n (%) | 27 (0.1%) | 8 (0.0%) |
| **Minimum alveolar concentration** | | |
| Measured value, median (25th, 75th) | 0.6 (0.4, 0.8) | 0.5 (0.3, 0.8) |
| Total measurements, n, (25th, 75th) | 140.0 (67.0, 227.0) | 150.0 (67.0, 238.0) |
| Frequency, n/hour, (25th, 75th) | 48.8 (39.0, 53.1) | 52.6 (42.9, 56.0) |
| Encounters missing, n (%) | 4868 (10.3%) | 2192 (10.8%) |

| | **Development Cohort** (6/1/2014 - 11/26/2018) | **Validation Cohort** (11/27/2018 - 9/20/2020) |
|---|---|---|
| **Core temperature** | | |
| Measured value, degrees Celsius, median (25th, 75th) | 36.8 (36.2, 37.3) | 36.8 (36.3, 37.4) |
| Total measurements, n, (25th, 75th) | 106.0 (22.0, 190.2) | 76.0 (1.0, 169.0) |
| Frequency, n/hour, (25th, 75th) | 38.8 (14.6, 46.3) | 32.7 (0.3, 44.0) |
| Encounters missing, n (%) | 3025 (6.4%) | 1758 (8.7%) |
| **Urine output sum, mL, median (25th, 75th)** | 160.0 (0.0, 500.0) | 100.0 (0.0, 515.0) |
| **Blood loss sum, mL, median (25th, 75th)** | 50.0 (0.0, 200.0) | 25.0 (0.0, 200.0) |

**Supplementary Table S5. Classification performance metrics for all experimental models and prediction phases in validation cohort.**

| | Sensitivity (95% CI) | Specificity (95% CI) | PPV (95% CI) | NPV (95% CI) | Accuracy (95% CI) | AUPRC (95% CI) | AUROC (95% CI) |
|---|---|---|---|---|---|---|---|
| **Prolonged ICU Stay** | | | | | | | |
| Baseline Preop | 0.81 (0.77-0.83) | 0.77 (0.75-0.81) | 0.64 (0.62-0.67) | 0.89 (0.87-0.90) | 0.78 (0.77-0.80) | 0.78 (0.76-0.79) | 0.87 (0.87-0.88) |
| Deep Preop | 0.80 (0.78-0.83) | 0.80 (0.77-0.82) | 0.67 (0.64-0.69) | 0.89 (0.88-0.90) | 0.80 (0.79-0.81) | 0.82 (0.81-0.82) | 0.89 (0.88-0.89) |
| Deep Preop (Multi-Task) | 0.81 (0.79-0.83) | 0.78 (0.76-0.80) | 0.65 (0.63-0.67) | 0.89 (0.88-0.90) | 0.79 (0.78-0.80) | 0.80 (0.79-0.81) | 0.88 (0.88-0.89) |
| Baseline Intraop | 0.79 (0.77-0.83) | 0.80 (0.76-0.82) | 0.66 (0.63-0.68) | 0.88 (0.88-0.90) | 0.80 (0.78-0.80) | 0.80 (0.79-0.81) | 0.88 (0.87-0.88) |
| Deep Intraop | 0.81 (0.79-0.83) | 0.82 (0.81-0.84) | 0.69 (0.68-0.72) | 0.90 (0.89-0.90) | 0.82 (0.81-0.83) | 0.83 (0.82-0.84) | 0.89 (0.88-0.89) |
| Deep Intraop (Multi-Task) | 0.77 (0.74-0.80) | 0.84 (0.82-0.87) | 0.71 (0.68-0.74) | 0.88 (0.87-0.89) | 0.82 (0.81-0.83) | 0.83 (0.82-0.84) | 0.88 (0.88-0.89) |
| Baseline Postop | 0.80 (0.78-0.82) | 0.83 (0.81-0.86) | 0.71 (0.69-0.73) | 0.89 (0.89-0.90) | 0.82 (0.82-0.83) | 0.83 (0.83-0.84) | 0.90 (0.89-0.90) |
| Deep Postop | 0.83 (0.80-0.85) | 0.84 (0.83-0.87) | 0.72 (0.71-0.76) | 0.91 (0.90-0.92) | 0.84 (0.83-0.85) | 0.87 (0.86-0.88) | 0.92 (0.91-0.92) |
| Deep Postop (Multi-Task) | 0.83 (0.81-0.86) | 0.83 (0.80-0.85) | 0.71 (0.68-0.73) | 0.91 (0.90-0.92) | 0.83 (0.82-0.84) | 0.86 (0.85-0.86) | 0.91 (0.91-0.92) |
| **Prolonged Mechanical Ventilation** | | | | | | | |
| Baseline Preop | 0.77 (0.75-0.81) | 0.79 (0.75-0.80) | 0.24 (0.21-0.25) | 0.98 (0.97-0.98) | 0.79 (0.76-0.80) | 0.46 (0.44-0.49) | 0.86 (0.85-0.87) |
| Deep Preop | 0.77 (0.75-0.84) | 0.85 (0.78-0.87) | 0.30 (0.24-0.32) | 0.98 (0.98-0.98) | 0.84 (0.79-0.86) | 0.52 (0.50-0.55) | 0.89 (0.88-0.90) |
| Deep Preop (Multi-Task) | 0.84 (0.80-0.87) | 0.80 (0.76-0.84) | 0.26 (0.23-0.29) | 0.98 (0.98-0.99) | 0.80 (0.77-0.84) | 0.53 (0.50-0.55) | 0.90 (0.89-0.90) |
| Baseline Intraop | 0.81 (0.77-0.85) | 0.85 (0.81-0.88) | 0.32 (0.27-0.36) | 0.98 (0.98-0.99) | 0.85 (0.82-0.88) | 0.55 (0.52-0.58) | 0.90 (0.89-0.91) |
| Deep Intraop | 0.84 (0.82-0.86) | 0.85 (0.83-0.86) | 0.32 (0.30-0.34) | 0.98 (0.98-0.99) | 0.85 (0.84-0.86) | 0.59 (0.57-0.62) | 0.92 (0.91-0.92) |
| Deep Intraop (Multi-Task) | 0.85 (0.80-0.86) | 0.84 (0.84-0.89) | 0.31 (0.30-0.38) | 0.99 (0.98-0.99) | 0.84 (0.84-0.88) | 0.59 (0.56-0.61) | 0.91 (0.91-0.92) |
| Baseline Postop | 0.82 (0.79-0.87) | 0.86 (0.81-0.88) | 0.33 (0.28-0.37) | 0.98 (0.98-0.99) | 0.86 (0.81-0.88) | 0.58 (0.55-0.60) | 0.91 (0.91-0.92) |
| Deep Postop | 0.84 (0.82-0.89) | 0.85 (0.80-0.87) | 0.31 (0.27-0.35) | 0.98 (0.98-0.99) | 0.85 (0.80-0.86) | 0.61 (0.59-0.64) | 0.92 (0.92-0.93) |
| Deep Postop (Multi-Task) | 0.87 (0.83-0.88) | 0.84 (0.84-0.88) | 0.32 (0.31-0.37) | 0.99 (0.98-0.99) | 0.85 (0.84-0.88) | 0.62 (0.59-0.65) | 0.93 (0.92-0.94) |
| **Wound Complications** | | | | | | | |
| Baseline Preop | 0.65 (0.59-0.69) | 0.68 (0.64-0.75) | 0.36 (0.34-0.38) | 0.88 (0.87-0.88) | 0.67 (0.65-0.71) | 0.45 (0.43-0.47) | 0.73 (0.72-0.73) |
| Deep Preop | 0.62 (0.60-0.75) | 0.76 (0.64-0.78) | 0.41 (0.36-0.43) | 0.88 (0.88-0.90) | 0.73 (0.66-0.74) | 0.52 (0.50-0.53) | 0.76 (0.75-0.77) |
| Deep Preop (Multi-Task) | 0.67 (0.65-0.77) | 0.72 (0.61-0.73) | 0.39 (0.35-0.41) | 0.89 (0.88-0.91) | 0.71 (0.65-0.72) | 0.53 (0.51-0.54) | 0.77 (0.76-0.78) |
| Baseline Intraop | 0.56 (0.49-0.73) | 0.55 (0.40-0.63) | 0.26 (0.24-0.27) | 0.82 (0.82-0.85) | 0.56 (0.47-0.60) | 0.27 (0.26-0.28) | 0.59 (0.58-0.60) |
| Deep Intraop | 0.40 (0.36-0.59) | 0.76 (0.57-0.81) | 0.31 (0.27-0.33) | 0.82 (0.82-0.84) | 0.68 (0.58-0.71) | 0.31 (0.30-0.33) | 0.61 (0.60-0.62) |
| Deep Intraop (Multi-Task) | 0.45 (0.38-0.61) | 0.74 (0.57-0.80) | 0.31 (0.28-0.35) | 0.83 (0.82-0.84) | 0.67 (0.58-0.71) | 0.33 (0.32-0.34) | 0.62 (0.61-0.63) |
| Baseline Postop | 0.63 (0.62-0.73) | 0.68 (0.57-0.68) | 0.35 (0.32-0.36) | 0.87 (0.87-0.89) | 0.67 (0.61-0.67) | 0.42 (0.40-0.43) | 0.71 (0.70-0.72) |
| Deep Postop | 0.69 (0.64-0.73) | 0.72 (0.68-0.77) | 0.40 (0.38-0.43) | 0.89 (0.89-0.90) | 0.71 (0.69-0.74) | 0.53 (0.52-0.55) | 0.78 (0.77-0.78) |
| Deep Postop (Multi-Task) | 0.74 (0.68-0.76) | 0.66 (0.65-0.72) | 0.38 (0.36-0.41) | 0.90 (0.89-0.91) | 0.68 (0.67-0.72) | 0.52 (0.51-0.54) | 0.77 (0.77-0.78) |
| **Neurological Complications** | | | | | | | |
| Baseline Preop | 0.79 (0.73-0.82) | 0.70 (0.67-0.76) | 0.40 (0.39-0.44) | 0.93 (0.92-0.94) | 0.72 (0.70-0.75) | 0.59 (0.57-0.60) | 0.83 (0.82-0.83) |
| Deep Preop | 0.77 (0.73-0.83) | 0.77 (0.71-0.81) | 0.46 (0.42-0.49) | 0.93 (0.92-0.94) | 0.77 (0.73-0.79) | 0.63 (0.62-0.65) | 0.85 (0.84-0.86) |
| Deep Preop (Multi-Task) | 0.79 (0.78-0.81) | 0.74 (0.73-0.76) | 0.44 (0.42-0.45) | 0.93 (0.93-0.94) | 0.75 (0.74-0.76) | 0.63 (0.61-0.64) | 0.85 (0.84-0.85) |
| Baseline Intraop | 0.76 (0.68-0.77) | 0.61 (0.60-0.69) | 0.33 (0.32-0.36) | 0.91 (0.89-0.91) | 0.64 (0.63-0.69) | 0.45 (0.44-0.47) | 0.75 (0.74-0.76) |
| Deep Intraop | 0.70 (0.65-0.73) | 0.67 (0.65-0.72) | 0.35 (0.34-0.37) | 0.90 (0.89-0.91) | 0.68 (0.66-0.70) | 0.45 (0.44-0.47) | 0.75 (0.74-0.76) |
| Deep Intraop (Multi-Task) | 0.71 (0.63-0.75) | 0.67 (0.63-0.75) | 0.35 (0.34-0.39) | 0.90 (0.89-0.91) | 0.68 (0.65-0.73) | 0.48 (0.46-0.50) | 0.76 (0.75-0.77) |
| Baseline Postop | 0.76 (0.70-0.82) | 0.70 (0.64-0.77) | 0.39 (0.37-0.43) | 0.92 (0.91-0.93) | 0.71 (0.68-0.75) | 0.54 (0.52-0.55) | 0.81 (0.80-0.81) |
| Deep Postop | 0.80 (0.75-0.82) | 0.75 (0.73-0.80) | 0.45 (0.43-0.49) | 0.94 (0.93-0.94) | 0.76 (0.75-0.79) | 0.64 (0.62-0.65) | 0.86 (0.85-0.86) |

| | Sensitivity (95% CI) | Specificity (95% CI) | PPV (95% CI) | NPV (95% CI) | Accuracy (95% CI) | AUPRC (95% CI) | AUROC (95% CI) |
|---|---|---|---|---|---|---|---|
| Deep Postop (Multi-Task) | 0.83 (0.76-0.84) | 0.71 (0.70-0.78) | 0.42 (0.41-0.47) | 0.94 (0.93-0.95) | 0.74 (0.73-0.77) | 0.64 (0.62-0.65) | 0.85 (0.85-0.86) |
| **Cardiovascular Complications** | | | | | | | |
| Baseline Preop | 0.74 (0.70-0.78) | 0.69 (0.65-0.73) | 0.32 (0.30-0.33) | 0.93 (0.93-0.94) | 0.70 (0.67-0.72) | 0.42 (0.40-0.44) | 0.78 (0.77-0.79) |
| Deep Preop | 0.72 (0.67-0.77) | 0.75 (0.70-0.80) | 0.36 (0.33-0.39) | 0.93 (0.93-0.94) | 0.74 (0.71-0.78) | 0.48 (0.46-0.50) | 0.81 (0.80-0.82) |
| Deep Preop (Multi-Task) | 0.72 (0.67-0.75) | 0.74 (0.71-0.78) | 0.35 (0.33-0.38) | 0.93 (0.92-0.94) | 0.74 (0.72-0.76) | 0.48 (0.46-0.50) | 0.81 (0.80-0.81) |
| Baseline Intraop | 0.72 (0.71-0.77) | 0.76 (0.71-0.76) | 0.37 (0.33-0.38) | 0.93 (0.93-0.94) | 0.75 (0.72-0.76) | 0.46 (0.45-0.48) | 0.81 (0.80-0.82) |
| Deep Intraop | 0.74 (0.71-0.78) | 0.78 (0.73-0.80) | 0.39 (0.36-0.42) | 0.94 (0.93-0.95) | 0.77 (0.74-0.79) | 0.53 (0.51-0.55) | 0.83 (0.82-0.84) |
| Deep Intraop (Multi-Task) | 0.76 (0.72-0.77) | 0.76 (0.76-0.80) | 0.38 (0.37-0.41) | 0.94 (0.94-0.95) | 0.76 (0.76-0.79) | 0.53 (0.51-0.55) | 0.83 (0.82-0.84) |
| Baseline Postop | 0.74 (0.73-0.81) | 0.77 (0.70-0.77) | 0.38 (0.34-0.39) | 0.94 (0.94-0.95) | 0.76 (0.72-0.77) | 0.48 (0.46-0.50) | 0.82 (0.81-0.83) |
| Deep Postop | 0.77 (0.74-0.82) | 0.77 (0.72-0.81) | 0.40 (0.36-0.43) | 0.95 (0.94-0.96) | 0.77 (0.74-0.80) | 0.56 (0.54-0.58) | 0.85 (0.84-0.86) |
| Deep Postop (Multi-Task) | 0.83 (0.77-0.84) | 0.72 (0.71-0.78) | 0.37 (0.36-0.40) | 0.96 (0.95-0.96) | 0.74 (0.73-0.78) | 0.56 (0.54-0.58) | 0.85 (0.85-0.86) |
| **Sepsis** | | | | | | | |
| Baseline Preop | 0.73 (0.68-0.77) | 0.81 (0.77-0.85) | 0.27 (0.24-0.32) | 0.97 (0.97-0.97) | 0.81 (0.77-0.84) | 0.45 (0.43-0.48) | 0.85 (0.84-0.86) |
| Deep Preop | 0.79 (0.73-0.82) | 0.78 (0.76-0.83) | 0.26 (0.24-0.30) | 0.97 (0.97-0.98) | 0.78 (0.76-0.82) | 0.48 (0.46-0.50) | 0.86 (0.86-0.87) |
| Deep Preop (Multi-Task) | 0.80 (0.75-0.83) | 0.78 (0.74-0.83) | 0.26 (0.24-0.29) | 0.98 (0.97-0.98) | 0.78 (0.75-0.82) | 0.48 (0.45-0.50) | 0.87 (0.86-0.88) |
| Baseline Intraop | 0.70 (0.66-0.72) | 0.72 (0.71-0.73) | 0.19 (0.18-0.21) | 0.96 (0.96-0.96) | 0.72 (0.71-0.73) | 0.30 (0.28-0.32) | 0.77 (0.75-0.78) |
| Deep Intraop | 0.74 (0.67-0.76) | 0.68 (0.67-0.74) | 0.18 (0.17-0.21) | 0.96 (0.96-0.97) | 0.69 (0.68-0.74) | 0.32 (0.30-0.34) | 0.78 (0.77-0.79) |
| Deep Intraop (Multi-Task) | 0.70 (0.65-0.75) | 0.76 (0.72-0.81) | 0.22 (0.20-0.25) | 0.96 (0.96-0.97) | 0.75 (0.72-0.80) | 0.34 (0.32-0.37) | 0.80 (0.78-0.81) |
| Baseline Postop | 0.75 (0.72-0.79) | 0.80 (0.76-0.83) | 0.26 (0.23-0.28) | 0.97 (0.97-0.97) | 0.79 (0.76-0.82) | 0.41 (0.38-0.43) | 0.85 (0.84-0.85) |
| Deep Postop | 0.80 (0.72-0.82) | 0.79 (0.78-0.87) | 0.27 (0.25-0.35) | 0.98 (0.97-0.98) | 0.79 (0.78-0.86) | 0.50 (0.47-0.52) | 0.87 (0.87-0.88) |
| Deep Postop (Multi-Task) | 0.78 (0.75-0.82) | 0.81 (0.78-0.84) | 0.29 (0.25-0.32) | 0.97 (0.97-0.98) | 0.81 (0.78-0.83) | 0.50 (0.48-0.52) | 0.88 (0.87-0.88) |
| **Acute Kidney Injury** | | | | | | | |
| Baseline Preop | 0.75 (0.71-0.81) | 0.70 (0.64-0.73) | 0.34 (0.31-0.36) | 0.93 (0.93-0.94) | 0.71 (0.67-0.73) | 0.45 (0.43-0.47) | 0.80 (0.79-0.80) |
| Deep Preop | 0.78 (0.75-0.79) | 0.71 (0.70-0.74) | 0.35 (0.34-0.37) | 0.94 (0.93-0.94) | 0.72 (0.71-0.74) | 0.51 (0.49-0.53) | 0.82 (0.81-0.82) |
| Deep Preop (Multi-Task) | 0.74 (0.72-0.79) | 0.74 (0.69-0.76) | 0.37 (0.34-0.39) | 0.93 (0.93-0.94) | 0.74 (0.71-0.76) | 0.51 (0.49-0.53) | 0.82 (0.81-0.83) |
| Baseline Intraop | 0.55 (0.53-0.61) | 0.76 (0.71-0.76) | 0.32 (0.29-0.33) | 0.89 (0.89-0.90) | 0.72 (0.69-0.73) | 0.36 (0.35-0.38) | 0.71 (0.70-0.72) |
| Deep Intraop | 0.67 (0.53-0.71) | 0.65 (0.62-0.79) | 0.28 (0.27-0.35) | 0.91 (0.89-0.91) | 0.66 (0.63-0.75) | 0.40 (0.38-0.42) | 0.73 (0.72-0.74) |
| Deep Intraop (Multi-Task) | 0.57 (0.50-0.67) | 0.77 (0.67-0.84) | 0.33 (0.29-0.38) | 0.90 (0.89-0.91) | 0.73 (0.67-0.78) | 0.41 (0.39-0.43) | 0.74 (0.73-0.74) |
| Baseline Postop | 0.73 (0.72-0.79) | 0.73 (0.67-0.74) | 0.36 (0.33-0.37) | 0.93 (0.93-0.94) | 0.73 (0.69-0.74) | 0.45 (0.43-0.47) | 0.80 (0.79-0.81) |
| Deep Postop | 0.74 (0.72-0.82) | 0.76 (0.68-0.78) | 0.39 (0.34-0.40) | 0.93 (0.93-0.95) | 0.76 (0.70-0.77) | 0.54 (0.52-0.55) | 0.83 (0.82-0.83) |
| Deep Postop (Multi-Task) | 0.74 (0.70-0.82) | 0.76 (0.68-0.79) | 0.38 (0.34-0.41) | 0.93 (0.93-0.95) | 0.75 (0.70-0.78) | 0.52 (0.51-0.54) | 0.82 (0.82-0.83) |
| **Venous Thromboembolism** | | | | | | | |
| Baseline Preop | 0.69 (0.60-0.77) | 0.73 (0.66-0.82) | 0.13 (0.11-0.16) | 0.98 (0.97-0.98) | 0.73 (0.67-0.81) | 0.17 (0.15-0.19) | 0.78 (0.76-0.79) |
| Deep Preop | 0.74 (0.70-0.83) | 0.76 (0.65-0.79) | 0.15 (0.12-0.16) | 0.98 (0.98-0.99) | 0.76 (0.66-0.79) | 0.22 (0.20-0.24) | 0.82 (0.81-0.83) |
| Deep Preop (Multi-Task) | 0.78 (0.71-0.83) | 0.72 (0.67-0.78) | 0.14 (0.13-0.16) | 0.98 (0.98-0.99) | 0.72 (0.68-0.78) | 0.26 (0.23-0.29) | 0.82 (0.81-0.83) |
| Baseline Intraop | 0.52 (0.50-0.64) | 0.74 (0.64-0.75) | 0.10 (0.09-0.11) | 0.96 (0.96-0.97) | 0.73 (0.63-0.73) | 0.12 (0.11-0.13) | 0.67 (0.66-0.69) |
| Deep Intraop | 0.59 (0.51-0.65) | 0.75 (0.69-0.82) | 0.12 (0.10-0.14) | 0.97 (0.97-0.97) | 0.74 (0.69-0.80) | 0.16 (0.14-0.18) | 0.71 (0.69-0.73) |
| Deep Intraop (Multi-Task) | 0.62 (0.54-0.71) | 0.73 (0.65-0.81) | 0.12 (0.10-0.15) | 0.97 (0.97-0.98) | 0.73 (0.65-0.80) | 0.17 (0.15-0.19) | 0.74 (0.72-0.75) |
| Baseline Postop | 0.73 (0.70-0.76) | 0.67 (0.66-0.68) | 0.11 (0.11-0.12) | 0.98 (0.97-0.98) | 0.67 (0.67-0.68) | 0.14 (0.13-0.16) | 0.75 (0.74-0.77) |
| Deep Postop | 0.78 (0.70-0.83) | 0.72 (0.68-0.78) | 0.14 (0.12-0.16) | 0.98 (0.98-0.99) | 0.72 (0.69-0.78) | 0.23 (0.21-0.26) | 0.82 (0.81-0.83) |
| Deep Postop (Multi-Task) | 0.80 (0.72-0.85) | 0.70 (0.66-0.78) | 0.13 (0.12-0.16) | 0.98 (0.98-0.99) | 0.70 (0.67-0.78) | 0.25 (0.22-0.27) | 0.83 (0.81-0.84) |
| **In-Hospital Mortality** | | | | | | | |

| | Sensitivity (95% CI) | Specificity (95% CI) | PPV (95% CI) | NPV (95% CI) | Accuracy (95% CI) | AUPRC (95% CI) | AUROC (95% CI) |
|---|---|---|---|---|---|---|---|
| Baseline Preop | 0.75 (0.70-0.80) | 0.82 (0.82-0.83) | 0.06 (0.06-0.07) | 1.00 (0.99-1.00) | 0.82 (0.81-0.83) | 0.12 (0.10-0.15) | 0.84 (0.82-0.86) |
| Deep Preop | 0.83 (0.80-0.92) | 0.81 (0.72-0.81) | 0.06 (0.05-0.07) | 1.00 (1.00-1.00) | 0.81 (0.73-0.81) | 0.17 (0.14-0.22) | 0.90 (0.88-0.91) |
| Deep Preop (Multi-Task) | 0.79 (0.76-0.88) | 0.84 (0.72-0.85) | 0.07 (0.05-0.08) | 1.00 (1.00-1.00) | 0.83 (0.72-0.85) | 0.17 (0.13-0.20) | 0.89 (0.88-0.91) |
| Baseline Intraop | 0.74 (0.64-0.79) | 0.72 (0.72-0.84) | 0.04 (0.04-0.06) | 0.99 (0.99-1.00) | 0.72 (0.72-0.84) | 0.08 (0.06-0.10) | 0.79 (0.77-0.82) |
| Deep Intraop | 0.85 (0.73-0.88) | 0.72 (0.71-0.85) | 0.05 (0.04-0.07) | 1.00 (0.99-1.00) | 0.73 (0.71-0.85) | 0.15 (0.12-0.19) | 0.86 (0.84-0.88) |
| Deep Intraop (Multi-Task) | 0.83 (0.73-0.87) | 0.76 (0.74-0.85) | 0.05 (0.05-0.08) | 1.00 (0.99-1.00) | 0.76 (0.74-0.85) | 0.18 (0.14-0.22) | 0.88 (0.86-0.89) |
| Baseline Postop | 0.72 (0.62-0.79) | 0.82 (0.81-0.89) | 0.06 (0.05-0.09) | 0.99 (0.99-1.00) | 0.82 (0.81-0.88) | 0.11 (0.09-0.13) | 0.83 (0.81-0.86) |
| Deep Postop | 0.91 (0.78-0.94) | 0.73 (0.72-0.85) | 0.05 (0.05-0.08) | 1.00 (1.00-1.00) | 0.73 (0.72-0.85) | 0.19 (0.15-0.24) | 0.90 (0.89-0.92) |
| Deep Postop (Multi-Task) | 0.88 (0.79-0.95) | 0.79 (0.74-0.90) | 0.06 (0.05-0.11) | 1.00 (1.00-1.00) | 0.79 (0.74-0.90) | 0.20 (0.16-0.25) | 0.92 (0.91-0.93) |

PPV: positive predictive value; NPV: negative predictive value; AUPRC: area under the precision-recall curve; AUROC: area under the receiver operating characteristic curve.

**Supplementary Table S6. Summary of input features and description of variable preprocessing.**

| Variable Name | Type | Data Source | Categories | Preprocessing |
|---|---|---|---|---|
| **Demographic Variables** | | | | |
| Age | Continuous | Derived | | Outlier adjustment [a], Feature scaling [c] |
| Gender | Binary | Raw | 2 | |
| Race | Nominal | Raw | 3 | Missing value imputation [b], One-hot encoding [d] |
| Body mass index | Continuous | Raw | | Outlier adjustment [a], Missing value imputation [b], Feature scaling [c] |
| Marital status | Nominal | Raw | 3 | One-hot encoding [d] |
| Ethnicity | Binary | Raw | 2 | Missing value imputation [b] |
| Language | Binary | Raw | 2 | Missing value imputation [b] |
| Smoking status | Nominal | Raw | 3 | One-hot encoding [d] |
| Insurance | Nominal | Raw | 4 | One-hot encoding [d] |
| **Patient Neighborhood Characteristics** | | | | |
| ZIP code | Nominal | Raw | 2108 | Missing value imputation [b], Embedding representation [f] |
| Rural area | Binary | Derived | 2 | Missing value imputation [b] |
| Total population | Continuous | Derived | | Outlier adjustment [a], Missing value imputation [b], Feature scaling [c] |
| Median income | Continuous | Derived | | Outlier adjustment [a], Missing value imputation [b], Feature scaling [c] |
| African-American population proportion | Continuous | Derived | | Outlier adjustment [a], Missing value imputation [b], Feature scaling [c] |
| Hispanic population proportion | Continuous | Derived | | Outlier adjustment [a], Missing value imputation [b], Feature scaling [c] |
| Poverty rate | Continuous | Derived | | Outlier adjustment [a], Missing value imputation [b], Feature scaling [c] |
| Distance to hospital | Continuous | Derived | | Outlier adjustment [a], Missing value imputation [b], Feature scaling [c] |
| **Admission Information** | | | | |
| Month of admission | Nominal | Raw | 12 | Cyclical embedding [g], Feature scaling [c] |
| Day of admission | Nominal | Raw | 7 | Cyclical embedding [g], Feature scaling [c] |
| Hour of admission | Nominal | Raw | 24 | Cyclical embedding [g], Feature scaling [c] |
| Admission source | Binary | Raw | 2 | |
| Admission type (emergent/elective) | Binary | Derived | 2 | |
| Admission type (medicine/surgery) | Binary | Derived | 2 | |
| Night admission | Binary | Derived | 2 | |
| **Surgical Procedure Information** | | | | |
| Primary procedure | Nominal | Derived | 3232 | Embedding representation [f] |
| Attending surgeon | Nominal | Raw | 370 | Embedding representation [f] |
| Surgery type | Nominal | Derived | 17 | Embedding representation [f] |
| Operating room | Nominal | Raw | 63 | Embedding representation [f] |
| Operating room type (trauma) | Binary | Derived | 2 | |
| Anesthesia type | Binary | Derived | 2 | |

| **Variable Name** | **Type** | **Data Source** | **Categories** | **Preprocessing** |
|---|---|---|---|---|
| Scheduled postoperative location | Binary | Derived | 2 | |
| Time from admission to surgery | Continuous | Derived | | Outlier adjustment [a], Feature scaling [c] |
| Surgeon specialty | Nominal | Derived | 48 | Embedding representation [f] |
| **Admission comorbidities** | | | | |
| Myocardial infarction | Binary | Derived | 2 | |
| Congestive heart failure | Binary | Derived | 2 | |
| Peripheral vascular disease | Binary | Derived | 2 | |
| Cerebrovascular disease | Binary | Derived | 2 | |
| Chronic pulmonary disease | Binary | Derived | 2 | |
| Metastatic carcinoma | Binary | Derived | 2 | |
| Cancer | Binary | Derived | 2 | |
| Liver disease | Binary | Derived | 2 | |
| Diabetes | Binary | Derived | 2 | |
| Hypertension | Binary | Derived | 2 | |
| Hypothyroidism | Binary | Derived | 2 | |
| Valvular disease | Binary | Derived | 2 | |
| Coagulopathy | Binary | Derived | 2 | |
| Obesity | Binary | Derived | 2 | |
| Weight loss | Binary | Derived | 2 | |
| Fluid/electrolyte disorders | Binary | Derived | 2 | |
| Chronic anemia | Binary | Derived | 2 | |
| Alcohol or drug abuse | Binary | Derived | 2 | |
| Depression | Binary | Derived | 2 | |
| Charlson comorbidity index | Continuous | Derived | | |
| Number of unique diagnosis codes | Continuous | Derived | | |
| Chronic kidney disease | Binary | Derived | 2 | |
| **Medications History** | | | | |
| ACE Inhibitors | Binary | Derived | 2 | |
| Aminoglycosides | Binary | Derived | 2 | |
| Antiemetics | Binary | Derived | 2 | |
| Aspirin | Binary | Derived | 2 | |
| Beta Blockers | Binary | Derived | 2 | |
| Bicarbonates | Binary | Derived | 2 | |
| Corticosteroids | Binary | Derived | 2 | |
| Diuretics | Binary | Derived | 2 | |
| NSAIDS | Binary | Derived | 2 | |
| Vasopressors/Inotropes | Binary | Derived | 2 | |
| Statins | Binary | Derived | 2 | |
| Vancomycin | Binary | Derived | 2 | |
| Nephrotoxic | Binary | Derived | 2 | |
| Total number of medications | Continuous | Derived | | |
| **Laboratory Results History** | | | | |
| Urea nitrogen/creatinine ratio | Continuous | Raw | | Outlier adjustment [a], Missing value imputation [b], Feature scaling [c] |
| Hemoglobin | Continuous | Raw | | Outlier adjustment [a], Missing value imputation [b], Feature scaling [c] |
| Leukocytes | Continuous | Raw | | Outlier adjustment [a], Missing value imputation [b], Feature scaling [c] |
| Erythrocytes | Continuous | Raw | | Outlier adjustment [a], Missing value imputation [b], Feature scaling [c] |

| Variable Name | Type | Data Source | Categories | Preprocessing |
|---|---|---|---|---|
| Hematocrit | Continuous | Raw | | Outlier adjustment [a], Missing value imputation [b], Feature scaling [c] |
| Erythrocyte mean corpuscular volume | Continuous | Raw | | Outlier adjustment [a], Missing value imputation [b], Feature scaling [c] |
| Erythrocyte mean corpuscular hemoglobin concentration | Continuous | Raw | | Outlier adjustment [a], Missing value imputation [b], Feature scaling [c] |
| Erythrocyte mean corpuscular hemoglobin | Continuous | Raw | | Outlier adjustment [a], Missing value imputation [b], Feature scaling [c] |
| Erythrocyte distribution width | Continuous | Raw | | Outlier adjustment [a], Missing value imputation [b], Feature scaling [c] |
| Platelets | Continuous | Raw | | Outlier adjustment [a], Missing value imputation [b], Feature scaling [c] |
| Platelet mean volume | Continuous | Raw | | Outlier adjustment [a], Missing value imputation [b], Feature scaling [c] |
| Neutrophils | Continuous | Raw | | Outlier adjustment [a], Missing value imputation [b], Feature scaling [c] |
| Glucose, serum | Continuous | Raw | | Outlier adjustment [a], Missing value imputation [b], Feature scaling [c] |
| Urea nitrogen, serum | Continuous | Raw | | Outlier adjustment [a], Missing value imputation [b], Feature scaling [c] |
| Creatinine, serum | Continuous | Raw | | Outlier adjustment [a], Missing value imputation [b], Feature scaling [c] |
| Sodium, serum | Continuous | Raw | | Outlier adjustment [a], Missing value imputation [b], Feature scaling [c] |
| Potassium, serum | Continuous | Raw | | Outlier adjustment [a], Missing value imputation [b], Feature scaling [c] |
| Chloride, serum | Continuous | Raw | | Outlier adjustment [a], Missing value imputation [b], Feature scaling [c] |
| Carbon dioxide, serum | Continuous | Raw | | Outlier adjustment [a], Missing value imputation [b], Feature scaling [c] |
| Lactate, serum | Continuous | Raw | | Outlier adjustment [a], Missing value imputation [b], Feature scaling [c] |
| Calcium, serum | Continuous | Raw | | Outlier adjustment [a], Missing value imputation [b], Feature scaling [c] |
| Anion gap, serum | Continuous | Raw | | Outlier adjustment [a], Missing value imputation [b], Feature scaling [c] |
| Alanine, serum | Continuous | Raw | | Outlier adjustment [a], Missing value imputation [b], Feature scaling [c] |
| Albumin, serum | Continuous | Raw | | Outlier adjustment [a], Missing value imputation [b], Feature scaling [c] |

| Variable Name | Type | Data Source | Categories | Preprocessing |
| --- | --- | --- | --- | --- |
| Asparate, serum | Continuous | Raw | | Outlier adjustment [a], Missing value imputation [b], Feature scaling [c] |
| Bilirubin, serum | Continuous | Raw | | Outlier adjustment [a], Missing value imputation [b], Feature scaling [c] |
| C-reactive protein, serum | Continuous | Raw | | Outlier adjustment [a], Missing value imputation [b], Feature scaling [c] |
| Prothrombin time, serum | Continuous | Raw | | Outlier adjustment [a], Missing value imputation [b], Feature scaling [c] |
| Erythrocyte sedimentation rate, serum | Continuous | Raw | | Outlier adjustment [a], Missing value imputation [b], Feature scaling [c] |
| Troponin I, serum | Continuous | Raw | | Outlier adjustment [a], Missing value imputation [b], Feature scaling [c] |
| Troponin T, serum | Continuous | Raw | | Outlier adjustment [a], Missing value imputation [b], Feature scaling [c] |
| Erythrocytes, urine | Nominal | Derived | 4 | Missing value imputation [b], One-hot encoding [d] |
| Protein, urine | Nominal | Derived | 4 | Missing value imputation [b], One-hot encoding [d] |
| Glucose, urine | Nominal | Derived | 4 | Missing value imputation [b], One-hot encoding [d] |
| Hemoglobin, urine | Nominal | Derived | 4 | Missing value imputation [b], One-hot encoding [d] |
| Reference creatinine | Continuous | Derived | | Outlier adjustment [a], Missing value imputation [b], Feature scaling [c] |
| Estimated glomerular filtration rate | Continuous | Derived | | Outlier adjustment [a], Missing value imputation [b], Feature scaling [c] |
| **Intraoperative Variables** | | | | |
| Systolic blood pressure | Continuous | Raw | | Temporal processing [i], Time series creation [j], Feature scaling [c], Baseline time series extraction [k] |
| Diastolic blood pressure | Continuous | Raw | | Temporal processing [i], Time series creation [j], Feature scaling [c], Baseline time series extraction [k] |
| Mean arterial pressure | Continuous | Raw | | Temporal processing [i], Time series creation [j], Feature scaling [c], Baseline time series extraction [k] |
| Heart rate | Continuous | Raw | | Temporal processing [i], Time series creation [j], Feature scaling [c], Baseline time series extraction [k] |
| Oxygen saturation (SpO2) | Continuous | Raw | | Temporal processing [i], Time series creation [j], Feature scaling [c], Baseline time series extraction [k] |
| Fraction of inspired oxygen (FiO2) | Continuous | Raw | | Temporal processing [i], Time series creation [j], Feature scaling [c], Baseline time series extraction [k] |

| Variable Name | Type | Data Source | Categories | Preprocessing |
|---|---|---|---|---|
| End-tidal carbon dioxide (EtCO2) | Continuous | Raw | | Temporal processing [i], Time series creation [j], Feature scaling [c], Baseline time series extraction [k] |
| Tidal volume | Continuous | Raw | | Temporal processing [i], Time series creation [j], Feature scaling [c], Baseline time series extraction [k] |
| Respiration rate | Continuous | Raw | | Temporal processing [i], Time series creation [j], Feature scaling [c], Baseline time series extraction [k] |
| Peak inspiratory pressure | Continuous | Raw | | Temporal processing [i], Time series creation [j], Feature scaling [c], Baseline time series extraction [k] |
| Minimum alveolar concentration | Continuous | Raw | | Temporal processing [i], Time series creation [j], Feature scaling [c], Baseline time series extraction [k] |
| Core temperature | Continuous | Raw | | Temporal processing [i], Time series creation [j], Feature scaling [c], Baseline time series extraction [k] |
| Urine output sum | Continuous | Raw | | Outlier adjustment [a], Missing value imputation [b], Feature scaling [c] |
| Blood loss sum | Continuous | Raw | | Outlier adjustment [a], Missing value imputation [b], Feature scaling [c] |
| Surgery duration | Continuous | Raw | | Outlier adjustment [a], Missing value imputation [b], Feature scaling [c] |

[a] For continuous variables, values that fell in the top and bottom 1% of its distribution were considered outliers and capped to the respective values given at the 1st and 99th percentiles.
[b] Missing numerical values were replaced with the median from the development cohort, and missing nominal variables were assigned to a distinct “missing” category.
[c] Continuous variables were standardized to zero mean and unit variance.
[d] Nominal variables with less than 10 levels were represented as zero vectors of length equal to the number of levels, with level indicators equal to one.
[e] Using residency zip code, we linked to US Census data to calculate residing neighborhood characteristics and distance from hospital.
[f] Nominal variables with 10 levels or greater were transformed to a numeric integer identifier ranging from 0 to the number of unique levels minus one, where implicit variable representations were learned as part of the model training process.
[g] To preserve relative proximity, temporally recurring features such as month and day of admission were cyclically embedded as two separate features by sine and cosine-based transformation. For example, December (12) is near January (1), and Sunday (7) is near Monday (1).
[h] Medications were taken within one year timeframe prior to surgery using RxNorms data grouped into drug classes according to the US, Department of Veterans Affairs National Drug File-Reference Terminology [24].
[i] Measurement values lying outside of expert-defined clinically normal value ranges for each variable were discarded. If two measurements existed at the same timestamp for a given patient, a random measurement was kept.
[j] For each surgical procedure, a time series was constructed by arranging intraoperative measurements chronologically, resampling to one-minute frequency intervals, performing linear interpolation in both directions (except for blood loss and urine output, which were imputed with zero), and imputing the development median at every timestep for procedures lacking a single measurement of a particular variable.
[k] For baseline models, a set of 49 statistical features was extracted from each intraoperative time series. This set included the following features: minimum, maximum, mean, median, standard deviation, sum of values, variance,

| Variable Name | Type | Data Source | Categories | Preprocessing |
|---|---|---|---|---|
| kurtosis, skewness, absolute energy, absolute sum of changes, counts above and below mean, first and last locations of both minimum and maximum, sequence length, longest strike above and below mean, mean absolute change, mean change, ratio of unique values to sequence length, variance larger than standard deviation, 9 quantiles, 9 index mass quantiles, 10-binned entropy, number of peaks, and range count. | | | | |

**Supplementary Table S7. Net reclassification index results.**

| Complication | Model | NRI | p | Event | Non-Event | Overall |
|---|---|---|---|---|---|---|
| Prolonged ICU Stay | Deep Learning (Multi-Task) | 0.07 (0.06-0.08) | <0.01 | 2.25 | 4.48 | 3.74 |
| | Deep Learning | 0.07 (0.06-0.08) | <0.01 | 3.49 | 3.67 | 3.61 |
| | Random Forest | 0.06 (0.04-0.07) | <0.01 | -0.49 | 6.11 | 3.91 |
| Prolonged Mechanical Ventilation | Deep Learning (Multi-Task) | 0.08 (0.06-0.10) | <0.01 | 2.92 | 4.97 | 4.81 |
| | Deep Learning | 0.07 (0.05-0.09) | <0.01 | 7.31 | -0.25 | 0.34 |
| | Random Forest | 0.12 (0.10-0.14) | <0.01 | 5.02 | 6.74 | 6.61 |
| Wound Complications | Deep Learning (Multi-Task) | 0.02 (0.00-0.03) | 0.07 | 7.25 | -5.67 | -2.90 |
| | Deep Learning | 0.03 (0.01-0.04) | <0.01 | 6.79 | -4.03 | -1.71 |
| | Random Forest | -0.02 (-0.04--0.01) | 0.05 | -1.77 | -0.36 | -0.66 |
| Neurological Complications | Deep Learning (Multi-Task) | 0.01 (-0.00-0.02) | 0.44 | 3.26 | -2.73 | -1.52 |
| | Deep Learning | 0.01 (-0.00-0.02) | 0.21 | 3.04 | -2.21 | -1.14 |
| | Random Forest | -0.03 (-0.04--0.01) | 0.01 | -3.09 | 0.49 | -0.23 |
| Cardiovascular Complications | Deep Learning (Multi-Task) | 0.09 (0.08-0.11) | <0.01 | 10.78 | -1.75 | 0.29 |
| | Deep Learning | 0.07 (0.06-0.09) | <0.01 | 5.03 | 2.40 | 2.83 |
| | Random Forest | 0.08 (0.06-0.09) | <0.01 | -0.27 | 7.90 | 6.57 |
| Sepsis | Deep Learning (Multi-Task) | 0.01 (-0.00-0.03) | 0.17 | -1.97 | 3.31 | 2.85 |
| | Deep Learning | 0.02 (0.00-0.03) | 0.07 | 0.85 | 0.95 | 0.94 |
| | Random Forest | -0.00 (-0.02-0.02) | 0.86 | 1.63 | -1.84 | -1.54 |
| Acute Kidney Injury | Deep Learning (Multi-Task) | 0.01 (-0.00-0.02) | 0.36 | -0.38 | 1.06 | 0.82 |
| | Deep Learning | 0.01 (-0.00-0.02) | 0.25 | -4.62 | 5.51 | 3.79 |
| | Random Forest | 0.01 (-0.00-0.03) | 0.22 | -1.69 | 3.01 | 2.21 |
| Venous Thromb. | Deep Learning (Multi-Task) | 0.01 (-0.01-0.03) | 0.68 | 2.63 | -2.06 | -1.81 |
| | Deep Learning | 0.00 (-0.02-0.02) | 0.94 | 3.81 | -3.71 | -3.30 |
| | Random Forest | -0.02 (-0.05-0.01) | 0.31 | 4.00 | -6.11 | -5.56 |
| In-Hospital Mortality | Deep Learning (Multi-Task) | 0.05 (0.01-0.09) | 0.07 | 9.35 | -4.57 | -4.35 |
| | Deep Learning | 0.00 (-0.04-0.05) | 0.92 | 8.41 | -8.13 | -7.87 |
| | Random Forest | -0.03 (-0.09-0.03) | 0.38 | -3.12 | -0.25 | -0.29 |

## Supplementary Table S8. Absolute and relative risk between preoperative and postoperative models.

| Outcome | Prevalence (%) | Preoperative | | | | | Postoperative | | | | |
|---|---|---|---|---|---|---|---|---|---|---|---|
| | | Low Risk | | High Risk | | Relative Risk (%) | Low Risk | | High Risk | | Relative Risk (%) (+/- Preoperative) |
| | | N (%) | Absolute Risk (%) | N (%) | Absolute Risk (%) | | N (%) | Absolute Risk (+/- Preoperative) | N (%) | Absolute Risk (+/- Preoperative) | |
| Prolonged ICU Stay | 33.3 | 11899 (58.6) | 10.9 | 8394 (41.4) | 65.2 | 6.0 | 12353 (60.9) | 9.2 (-1.6) | 7940 (39.1) | 70.8 (+5.6) | 7.7 (+1.7) |
| Prolonged Mechanical Ventilation | 7.8 | 15140 (74.6) | 1.7 | 5153 (25.4) | 25.6 | 15.1 | 16025 (79.0) | 1.3 (-0.4) | 4268 (21.0) | 32.0 (+6.4) | 24.4 (+9.3) |
| Wound Complications | 21.4 | 12932 (63.7) | 11.2 | 7361 (36.3) | 39.5 | 3.5 | 11713 (57.7) | 9.6 (-1.5) | 8580 (42.3) | 37.5 (-1.9) | 3.9 (+0.4) |
| Neurological Complications | 20.2 | 12808 (63.1) | 6.6 | 7485 (36.9) | 43.6 | 6.6 | 12232 (60.3) | 5.8 (-0.8) | 8061 (39.7) | 42.2 (-1.5) | 7.3 (+0.6) |
| Cardiovasc. Complications | 16.3 | 13505 (66.6) | 6.9 | 6788 (33.4) | 34.9 | 5.1 | 12851 (63.3) | 4.5 (-2.4) | 7442 (36.7) | 36.6 (+1.7) | 8.2 (+3.1) |
| Sepsis | 8.7 | 14823 (73.0) | 2.4 | 5470 (27.0) | 25.9 | 10.7 | 15471 (76.2) | 2.6 (+0.1) | 4822 (23.8) | 28.6 (+2.8) | 11.2 (+0.6) |
| Acute Kidney Injury | 16.9 | 13442 (66.2) | 6.6 | 6851 (33.8) | 37.2 | 5.6 | 13634 (67.2) | 6.6 (0.0) | 6659 (32.8) | 38.1 (+0.9) | 5.7 (+0.1) |
| Venous Thromb. | 5.4 | 14029 (69.1) | 1.7 | 6264 (30.9) | 13.7 | 7.9 | 13604 (67.0) | 1.6 (-0.2) | 6689 (33.0) | 13.2 (-0.4) | 8.4 (+0.5) |
| In-Hospital Mortality | 1.6 | 16754 (82.6) | 0.4 | 3539 (17.4) | 7.1 | 17.6 | 15812 (77.9) | 0.2 (-0.2) | 4481 (22.1) | 6.3 (-0.8) | 26.3 (+8.7) |

**Supplementary Table S9. Risk group transitions and corresponding inpatient mortality rate.**

| Outcome | Preoperative Risk Group | Postoperative Risk Group | N | % Encounters | % Preoperative Risk Group | Transition Group Mortality Rate |
|---|---|---|---|---|---|---|
| **Prolonged ICU Stay** | High | High | 6705 | 33.0 | 79.9 | 4.4 |
| | High | Low | 1689 | 8.3 | 20.1 | 0.5 |
| | Low | High | 1235 | 6.1 | 10.4 | 0.6 |
| | Low | Low | 10664 | 52.6 | 89.6 | 0.1 |
| **Prolonged Mechanical Ventilation** | High | High | 3514 | 17.3 | 68.2 | 7.3 |
| | High | Low | 1639 | 8.1 | 31.8 | 0.9 |
| | Low | High | 754 | 3.7 | 5.0 | 2.0 |
| | Low | Low | 14386 | 70.9 | 95.0 | 0.3 |
| **Wound Complications** | High | High | 6381 | 31.4 | 86.7 | 3.5 |
| | High | Low | 980 | 4.8 | 13.3 | 0.8 |
| | Low | High | 2199 | 10.8 | 17.0 | 1.5 |
| | Low | Low | 10733 | 52.9 | 83.0 | 0.5 |
| **Neurological Complications** | High | High | 6774 | 33.4 | 90.5 | 4.3 |
| | High | Low | 711 | 3.5 | 9.5 | 0.1 |
| | Low | High | 1287 | 6.3 | 10.0 | 1.4 |
| | Low | Low | 11521 | 56.8 | 90.0 | 0.1 |
| **Cardiovascular Complications** | High | High | 5381 | 26.5 | 79.3 | 5.2 |
| | High | Low | 1407 | 6.9 | 20.7 | 0.6 |
| | Low | High | 2061 | 10.2 | 15.3 | 0.5 |
| | Low | Low | 11444 | 56.4 | 84.7 | 0.2 |
| **Sepsis** | High | High | 4269 | 21.0 | 78.0 | 5.5 |
| | High | Low | 1201 | 5.9 | 22.0 | 1.6 |
| | Low | High | 553 | 2.7 | 3.7 | 2.5 |
| | Low | Low | 14270 | 70.3 | 96.3 | 0.4 |
| **Acute Kidney Injury** | High | High | 5792 | 28.5 | 84.5 | 4.6 |
| | High | Low | 1059 | 5.2 | 15.5 | 0.8 |
| | Low | High | 867 | 4.3 | 6.4 | 1.6 |
| | Low | Low | 12575 | 62.0 | 93.6 | 0.2 |
| **Venous Thromboembolism** | High | High | 5431 | 26.8 | 86.7 | 5.0 |
| | High | Low | 833 | 4.1 | 13.3 | 1.0 |
| | Low | High | 1258 | 6.2 | 9.0 | 1.2 |
| | Low | Low | 12771 | 62.9 | 91.0 | 0.2 |
| **In-Hospital Mortality** | High | High | 3041 | 15.0 | 85.9 | 8.2 |
| | High | Low | 498 | 2.5 | 14.1 | 1.0 |
| | Low | High | 1440 | 7.1 | 8.6 | 2.4 |
| | Low | Low | 15314 | 75.5 | 91.4 | 0.2 |

**Supplementary Table S10. Risk transitions categorized by multi-task complication group.**

| Preoperative | | Postoperative | | Risk Category Transition (Preoperative to Postoperative) | | | | | |
|---|---|---|---|---|---|---|---|---|---|
| Risk Category | N (%) | Risk Category | N (%) | Preoperative Risk Category | Postoperative Risk Category | N | % Encounters | % Preoperative Risk Category | Transition Group Mortality (%) |
| All High | 2275 (11.2) | All High | 2397 (11.8) | All High | All High | 1709 | 8.4 | 75.1 | 10.6 |
| | | | | All High | Mixed | 566 | 2.8 | 24.9 | 2.1 |
| | | | | All High | All Low | 0 | 0.0 | 0.0 | 0.0 |
| All Low | 6748 (33.3) | All Low | 6808 (33.5) | All Low | All High | 1 | 0.0 | 0.0 | 0.0 |
| | | | | All Low | Mixed | 1218 | 6.0 | 18.0 | 0.2 |
| | | | | All Low | All Low | 5529 | 27.2 | 81.9 | 0.0 |
| Mixed | 11270 (55.5) | Mixed | 11088 (54.6) | Mixed | All High | 687 | 3.4 | 6.1 | 4.9 |
| | | | | Mixed | Mixed | 9304 | 45.8 | 82.6 | 0.9 |
| | | | | Mixed | All Low | 1279 | 6.3 | 11.3 | 0.1 |

## SUPPLEMENTARY FIGURES

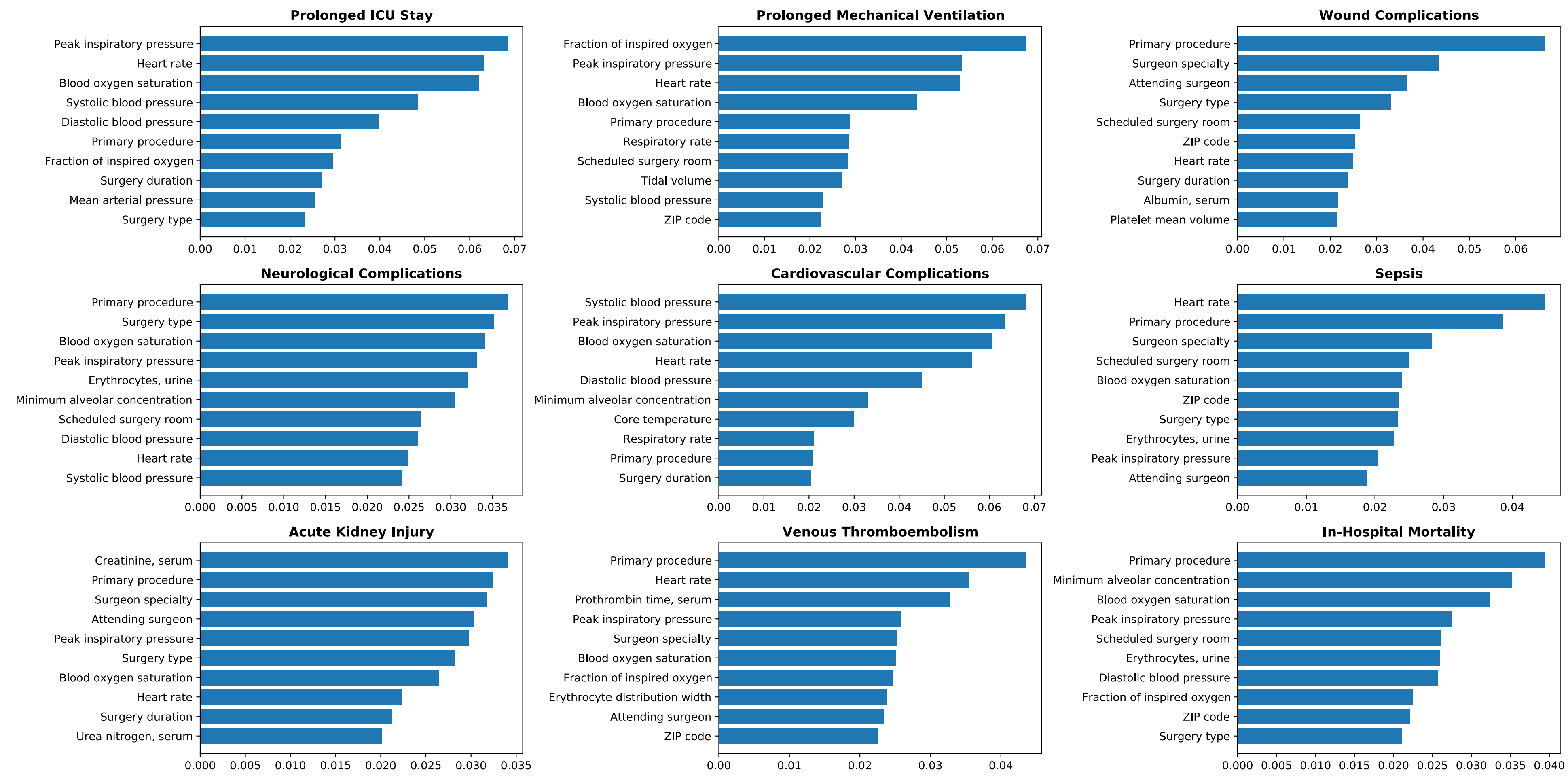

**Supplementary Figure S1. Ten features with largest integrated gradients attributions across validation cohort.**

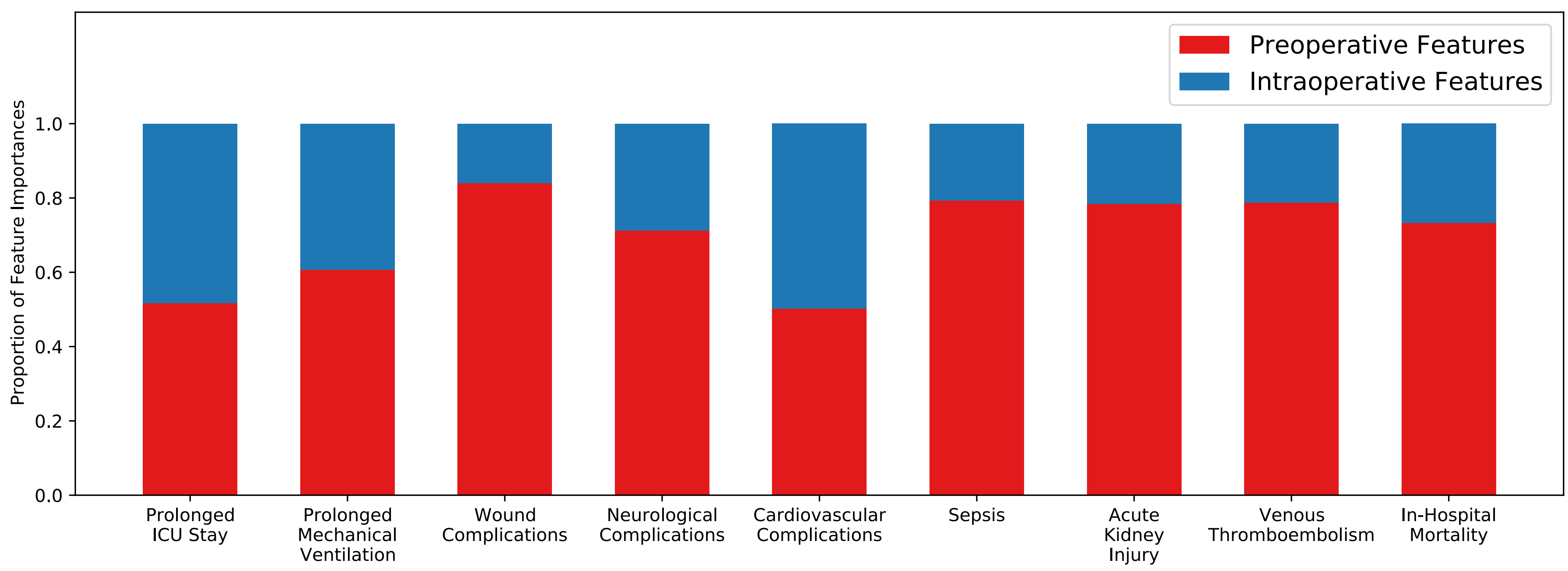

**Supplementary Figure S2. Distribution of integrated gradients feature attributions in validation cohort grouped by preoperative and intraoperative feature types.**

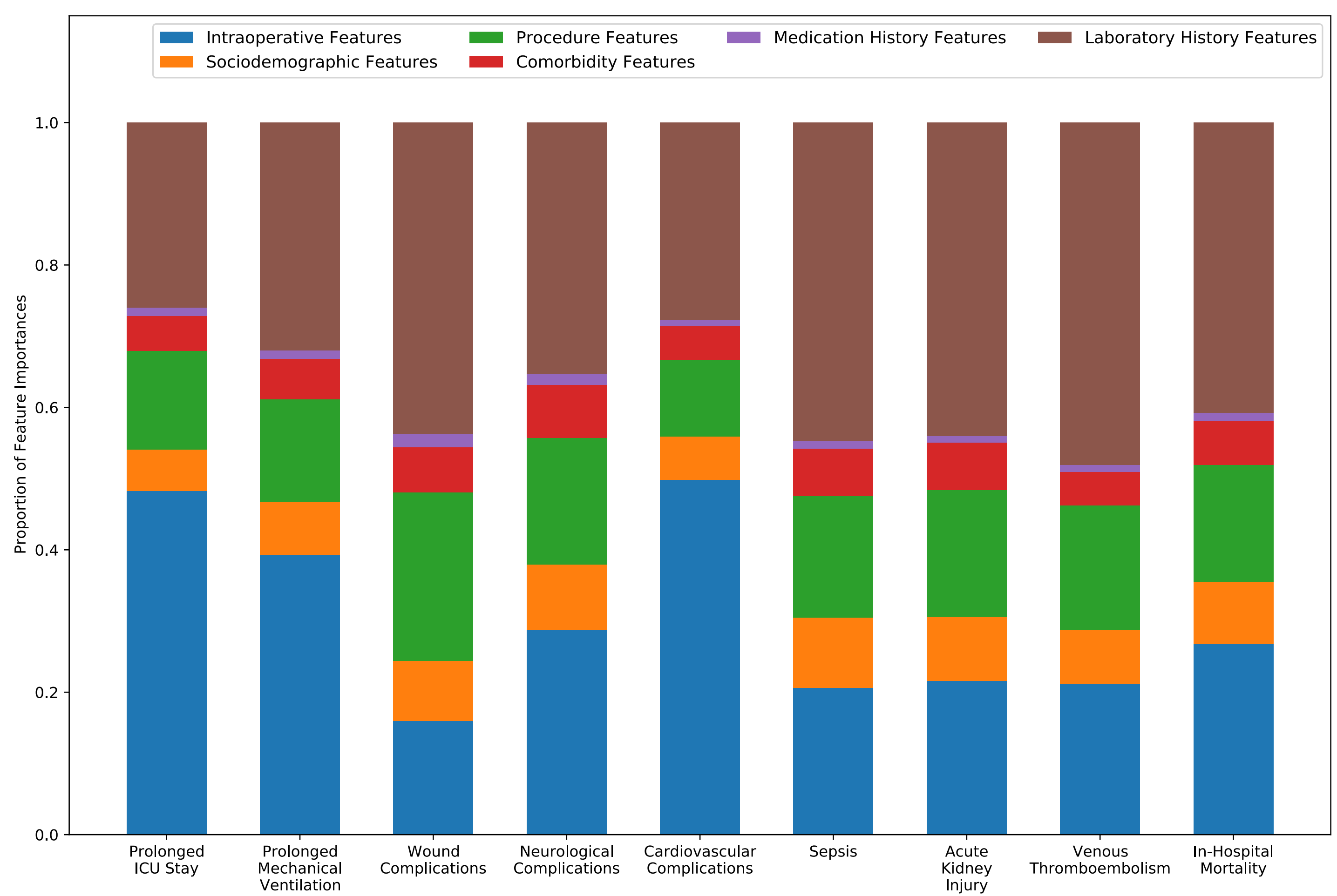

**Supplementary Figure S3. Distribution of integrated gradients feature attributions in validation cohort grouped by input variable type.**

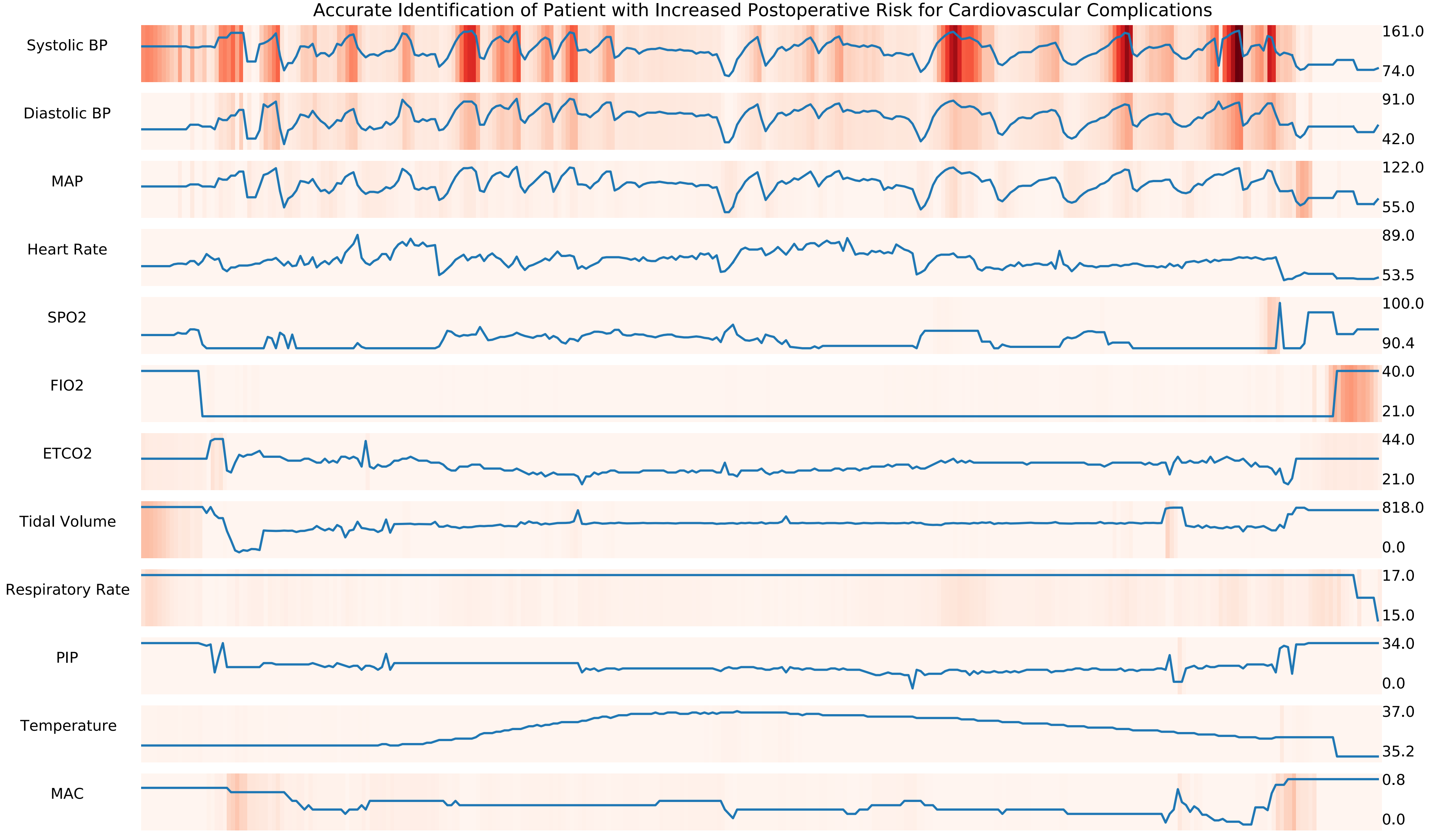

**Supplementary Figure S4. Temporal integrated gradients attributions for an example patient developing postoperative cardiovascular complications. The model correctly predicted elevated risk based on intraoperative physiological time series.**

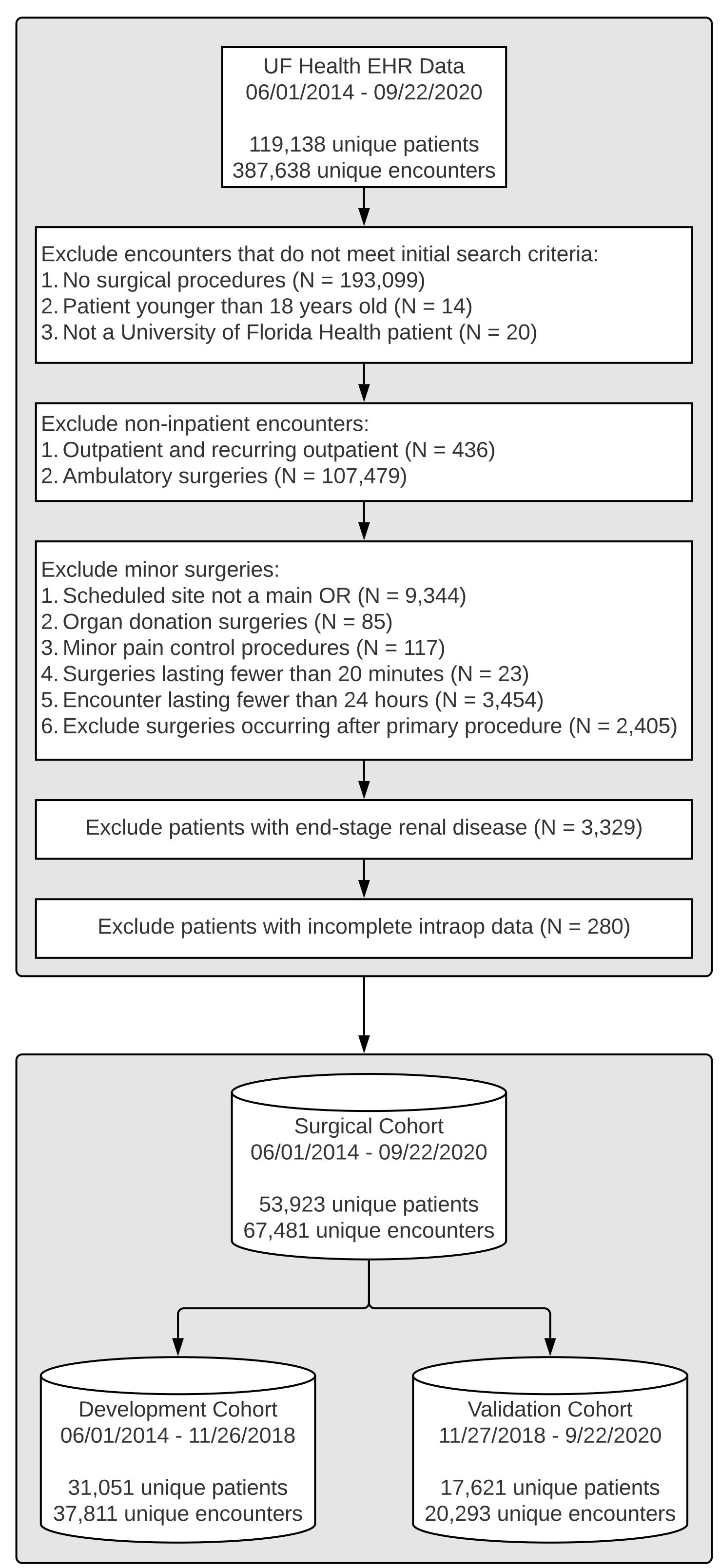

**Supplementary Figure S5. Overview of cohort selection criteria and derivation of development and validation cohort.**

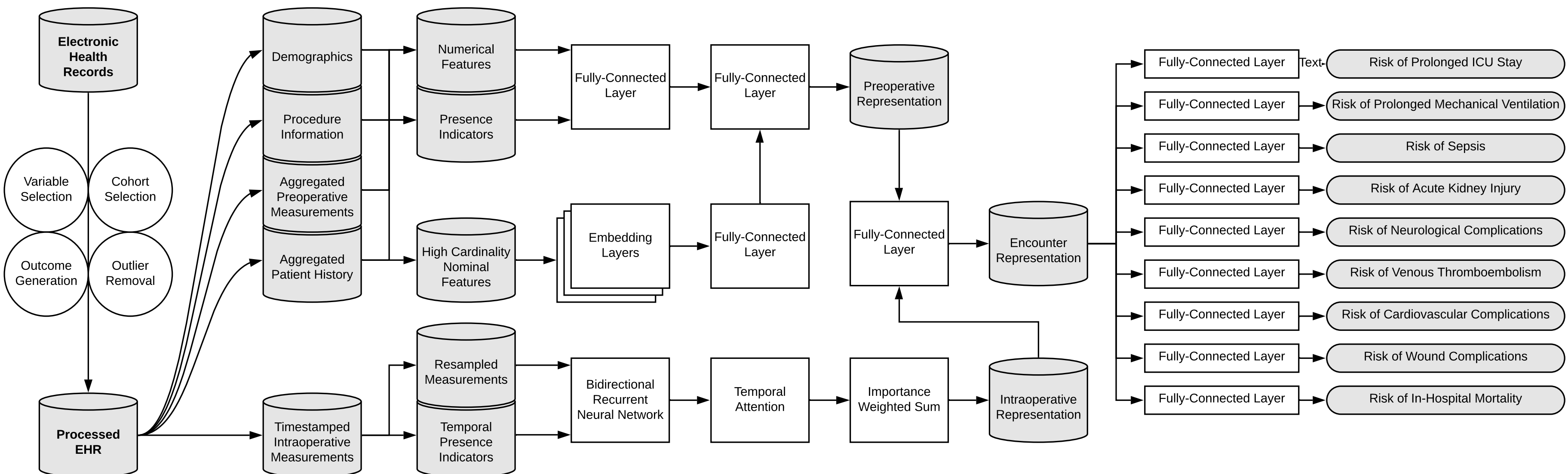

**Supplementary Figure S6. Data pipeline and neural network architecture.**

## SUPPLEMENTARY REFERENCES